\pdfoutput=1

\documentclass[11pt]{article}

\usepackage[final]{acl}

\usepackage{times}
\usepackage{latexsym}

\usepackage[T1]{fontenc}

\usepackage[utf8]{inputenc}

\usepackage{microtype}

\usepackage{inconsolata}

\usepackage{graphicx}

\usepackage[utf8]{inputenc} 
\usepackage[T1]{fontenc}    
\usepackage{hyperref}       
\usepackage{url}            
\usepackage{booktabs}       
\usepackage{amsfonts}       
\usepackage{nicefrac}       
\usepackage{microtype}      
\usepackage{xcolor}         

\usepackage{hyperref}
\usepackage{url}
\usepackage[utf8]{inputenc} %
\usepackage[T1]{fontenc}    %
\usepackage{hyperref}       %
\usepackage{url}            %
\usepackage{colortbl}
\usepackage{soul}
\usepackage{amsfonts}       %
\usepackage{microtype}
\usepackage{paralist}
\usepackage{amsmath}
\usepackage{caption}
\usepackage{booktabs}  
\usepackage{graphicx}
\usepackage{multirow}
\usepackage{subcaption}
\usepackage{diagbox}
\usepackage{bbding}
\usepackage{pifont}
\usepackage{lipsum}
\usepackage{capt-of} 
\usepackage{tabularx}
\usepackage{makecell}
\usepackage{varwidth}
\usepackage{pdfpages}
\usepackage{times}
\usepackage{etoc} %
\usepackage{array, makecell}
\usepackage[export]{adjustbox}
\usepackage{xcolor}
\usepackage{longtable}
\usepackage{float}
\usepackage{pifont}
\usepackage{amssymb}  
\usepackage{listings}
\usepackage{enumitem}
\usepackage{algorithm}
\usepackage{algpseudocode}
\usepackage[skins]{tcolorbox}
\usepackage[detect-all]{siunitx}

\usepackage{color}
\usepackage{multirow}
\usepackage{array}
\usepackage{booktabs}
\usepackage{enumitem}

\long\def\comment#1{}

\newcommand{\etc}{\textit{etc}}

\makeatletter
\newcommand{\printfnsymbol}[1]{%
  \textsuperscript{\@fnsymbol{#1}}%
}
\makeatother

\makeatletter
\def\@fnsymbol#1{\ensuremath{\ifcase#1\or \dagger\or \ddagger\or
   \mathsection\or \mathparagraph\or \|\or **\or \dagger\dagger
   \or \ddagger\ddagger \else\@ctrerr\fi}}
\makeatother

\title{Benchmarking Open-ended Audio Dialogue Understanding \\ for Large Audio-Language Models}

\author{Kuofeng Gao\textsuperscript{\rm 1}, Shu-Tao Xia\textsuperscript{\rm 1,3}\thanks{Corresponding authors.}, Ke Xu\textsuperscript{\rm 1}, Philip Torr\textsuperscript{\rm 2}, Jindong Gu\textsuperscript{\rm 2}\printfnsymbol{1}\\
\textsuperscript{\rm 1}Tsinghua University, \textsuperscript{\rm 2} University of Oxford, \textsuperscript{\rm 3} Peng Cheng Laboratory\\
\texttt{gkf21@mails.tsinghua.edu.cn, xiast@sz.tsinghua.edu.cn}\\
\texttt{xuke@tsinghua.edu.cn, \{philip.torr,jindong.gu\}@eng.ox.ac.uk}
}

\begin{document}
\maketitle
\begin{abstract}
Large Audio-Language Models (LALMs), such as GPT-4o, have recently unlocked audio dialogue capabilities, enabling direct spoken exchanges with humans. The potential of LALMs broadens their applicability across a wide range of practical scenarios supported by audio dialogues. However, given these advancements, a comprehensive benchmark to evaluate the performance of LALMs in the open-ended audio dialogue understanding remains absent currently. To address this gap, we propose an \textbf{A}udio \textbf{D}ialogue \textbf{U}nderstanding \textbf{Bench}mark \textbf{(ADU-Bench)}, which consists of 4 benchmark datasets. They assess the open-ended audio dialogue ability for LALMs in 3 general scenarios, 12 skills, 9 multilingual languages, and 4 categories of ambiguity handling. 
Notably, \textit{we firstly propose the evaluation of ambiguity handling} in audio dialogues that expresses different intentions beyond the same literal meaning of sentences, \textit{e.g.}, ``\texttt{Really!?}'' with different intonations. In summary, ADU-Bench includes over 20,000 open-ended audio dialogues for the assessment of LALMs. Through extensive experiments on 16 LALMs, our analysis reveals that existing LALMs struggle with mathematical symbols and formulas, understanding human behavior such as roleplay, comprehending multiple languages, and handling audio dialogue ambiguities from different phonetic elements, such as intonations, pause positions, and homophones. The benchmark is available at \url{https://adu-bench.github.io/}.
\end{abstract}

\section{Introduction}

Large Audio-Language Models (LALMs) \citep{chu2023qwen,tang2023salmonn,wu2023decoder,kong2024audio,lin2024advancing,xie2024mini,fu2024vita} have received attention for their abilities to handle various audio-related tasks. In particular, LALMs recently unlock unprecedented capabilities for interactive audio dialogues with humans. These dialogues are defined as a direct exchange of spoken language between LALMs and humans, which fosters a more dynamic mode of communication.
Recent advances, such as GPT-4o \citep{openaigpt4o}, have enabled LALMs to engage in back-and-forth dialogues with humans
and can observe various audio characteristics, which
broadens their applicability across diverse real-world situations that rely on interactive audio dialogues.


\begin{figure*}[t] \centering     
\includegraphics[width=\textwidth]{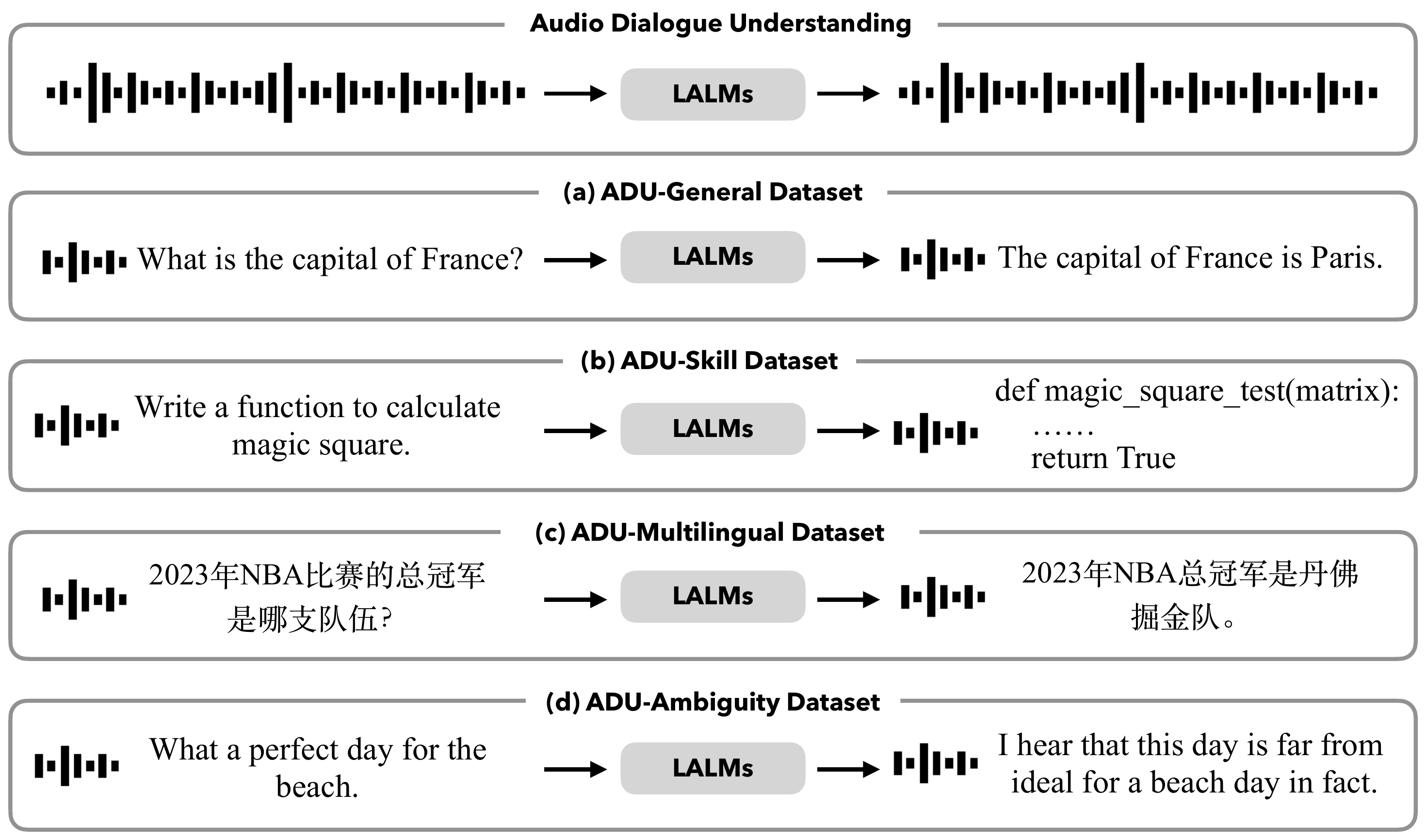} 
\vspace{-2em}
\caption{
ADU-Bench evaluates the open-ended audio dialogue understanding for LALMs, where users interact with LALMs directly through audio. Our ADU-Bench consists of 4 datasets, including (a) ADU-General dataset, (b) ADU-Skill dataset, (c) ADU-Multilingual dataset, and (d) ADU-Ambiguity dataset. In total, it encompasses 20,715 open-ended audio dialogues, comprising over 8,000 real-world recordings alongside synthetic audio samples.}
\label{fig:exmaple}
\end{figure*}

However, given these advancements, there is currently no comprehensive benchmark to evaluate LALMs' performance in handling open-ended audio dialogue understanding. Previous benchmarks on LALMs predominantly focus on their performance in multiple fundamental tasks~\citep{ huang2024dynamic,huang2024dynamic2}, audio question answering with text-based instructions~\citep{yang2024air,deshmukh2024audio,sakshi2024mmau,wang2025audiobench} or audio dialogues in general scenarios~\citep{ao2024sdeval,chen2024voicebench}. 
The absence of a comprehensive benchmark for evaluating LALMs in open-ended audio dialogues has led to suboptimal comparisons between different LALMs. 

Open-ended audio dialogues, where users can directly engage with LALMs through audio, constitute a significant portion of real-world interactions. These dialogues can encompass many 
subjects, such as helpful and daily questions, domain-specific skills, and multiple different languages.
Additionally, the variations in intonations or pause positions can allow speakers to express different intentions beyond the same literal meaning of sentences, adding further complexity to the dialogues. 
Therefore, the ability to handle open-ended audio dialogues effectively is crucial for LALMs to be truly useful in real-world applications. 

In this work, we propose an \textbf{A}udio \textbf{D}ialogue \textbf{U}nderstanding \textbf{Bench}mark \textbf{(ADU-Bench)}, a benchmark to evaluate the open-ended audio dialogue understanding for LALMs, which comprises 4 benchmark datasets as follows.
(1) The ADU-General dataset assesses the general dialogue understanding of LALMs, including 3 scenarios, \textit{i.e.}, helpful questions to query search engines, daily questions happening among human dialogues, and daily statements without rich contexts.  (2) The ADU-Skill dataset evaluates the skill-based dialogue ability, encompassing 12 different skills such as mathematics, physics, coding, \etc. (3) The ADU-Multilingual dataset tests the multilingual dialogue understanding, covering 9 languages, including English, French, and Chinese, \etc. (4) The ADU-Ambiguity dataset is designed to evaluate the audio dialogue ambiguity handling ability from different phonetic elements, including intonation-based, pause-based, homophone-based, and repetition-based ambiguity.
Notably, \textit{we firstly analyze the ambiguity within audio dialogues,} specifically addressing the challenge of different intentions that share the same literal sentence, such as the word ``\texttt{Really!?}'' spoken with different intonations.
In total, ADU-Bench comprises over 20,000 open-ended audio dialogues for LALMs.
An overall example of ADU-Bench is shown in Fig. \ref{fig:exmaple}.
For the evaluation, LALMs are first queried with user audio inputs and generate corresponding textual responses directly or convert audio responses into a textual format. Then, we primarily use GPT-4 \citep{achiam2023gpt} or manual annotation to generate references (expected ground truths) based on the textual transcriptions of each audio. 
Subsequently, following \citet{zheng2024judging,bai2024mt,yang2024air}, we include the transcriptions of audio, references, and responses into an evaluation prompt and use this prompt to query GPT-4 \citep{achiam2023gpt}, which generates a score for evaluating the quality of generated responses. However, the order in which the references and responses are presented in the evaluation prompt can influence the scores generated by GPT-4, leading to position bias \citep{zheng2024judging}. To eliminate position bias, we  conduct a second scoring by swapping the positions of the references and responses during evaluation. In addition, to eliminate bias from the GPT-4 based evaluation, we have included more LLMs for evaluation, such as LLaMA-3-70B-Instruct \citep{metallama3} and Qwen-2-72B-Instruct \citep{chu2023qwen}. 

We benchmark 16 popular LALMs on our ADU-Bench and analyze the results. Our analysis reveals: (1) There is still considerable room for improvement in the audio dialogue understanding of existing open-sourced LALMs. (2) LALMs face challenges when dealing with skills, such as Mathematics and Coding, which involve mathematical symbols and formulas. (3) LALMs exhibit limitations in handling tasks related to Common Sense and Roleplay, as they lack a deeper understanding of human behavior. (4) Existing LALMs struggle to comprehend different meanings of audio dialogues that have the same transcriptions, but differ in phonetic elements, such as intonations, pause positions, and homophones. 
We include some demonstrations of our audio dialogues on our project page.



\section{Related Work}
\textbf{Large Audio-language Models}.
Large audio-language models (LALMs)~\citep{chu2023qwen,tang2023salmonn,wu2023decoder,kong2024audio}  typically integrate audio modalities into large language models (LLMs)~\citep{touvron2023llama,openaigpt4o,gao2024denial,gao2024embedding,gao2024inducing,fang2025your,zou2025making,Kong2025WolfHI} to extend their capabilities for general-purpose audio and language understanding. LALMs can be broadly classified into two types: end-to-end LALMs and cascaded LALMs.
End-to-end LALMs can be further divided into two categories: (1) End-to-end LALMs specialize in audio understanding, which focus on integrating audio modality into LLMs, such as SpeechGPT \citep{zhang2023speechgpt}, BLSP \citep{wang2023blsp}, SALMONN \citep{tang2023salmonn}, and Qwen-Audio \citep{chu2023qwen}. (2) End-to-end LALMs extend their capabilities beyond audio understanding, which align various modalities into a single LLM, such as PandaGPT \citep{su2023pandagpt} and NExT-GPT \citep{wu2023next}.
Another approach involves cascaded LALMs like the combination of an automatic speech recognition model, such as Whisper-large \citep{radford2023robust}, and an LLM, such as GPT-4 \citep{achiam2023gpt}, to process a wide range of audio types. Our ADU-Bench aims to evaluate their performance in audio dialogue understanding across different domains.


\noindent \textbf{Benchmarks for LALMs}. 
Existing benchmarks for audio-related tasks can be broadly categorized into three areas: (1) fundamental audio tasks, (2) audio question answering with text-based instructions, and (3) audio dialogues.
For benchmarks focusing on fundamental audio tasks~\citep{huang2024dynamic,huang2024dynamic2}, evaluations are typically centered around specific objectives such as speech-to-text translation or emotion recognition.  In audio question answering with text-based instructions~\citep{yang2024air,deshmukh2024audio,sakshi2024mmau,wang2025audiobench}, models are required to interpret input audio and respond to input text-based instructions. 
In contrast, benchmarks for audio dialogues evaluate models to directly respond to audio queries without text-based instructions. While several established benchmarks~\citep{ao2024sdeval,chen2024voicebench} exist for audio dialogues, they predominantly focus on general scenarios, leaving a comprehensive benchmark unexplored. To bridge this gap, we propose ADU-Bench, which concentrates on evaluating LALMs in a wide range of audio dialogue scenarios.

\section{ADU-Bench}
\label{sec: Data Collection and Statistics}

\begin{table*}[t]
\caption{Data collection and statistics on 4 datasets in ADU-Bench, including dataset domains, dataset source, and dataset number. In total, ADU-Bench consists of 20,715 open-ended audio dialogues.
}
\vspace{-1em}
\small
\label{tab:statistics}
\centering
\setlength{\tabcolsep}{18.5pt}{
\begin{tabular}{lcccc}
\toprule
Datasets & Domains & Source & Number \\
\midrule
\multicolumn{1}{l}{\multirow{3}{*}{ADU-General}} & Helpful Question & Alpaca, NQ-Bench & \multicolumn{1}{c}{\multirow{3}{*}{12,000}} \\
& Daily Question & WebGLM, Slue HVB & \\
& Daily Statement & Common Voice & \\
\midrule
\multicolumn{1}{l}{\multirow{5}{*}{ADU-Skill}}  & Mathematics, Physics & \multicolumn{1}{c}{\multirow{5}{*}{\begin{tabular}[c]{@{}c@{}}GSM8K, MATH \\ WizardLM, ShareGPT \\ MBPP, MMLU \\ HotpotQA, StrategyQA \end{tabular}}}  & \multicolumn{1}{c}{\multirow{5}{*}{3,725}}  \\
& Chemistry, Biology & & \\
& Computer Science, Code, Law &   & \\
& Finance, Common Sense &  &  \\
&  Writing, Roleplay, Medicine &  &  \\
\midrule
\multicolumn{1}{l}{\multirow{3}{*}{ADU-Multilingual}} & Arabic, Chinese, English & \multicolumn{1}{c}{\multirow{3}{*}{\begin{tabular}[c]{@{}c@{}}Alpaca, NQ-Bench \\ WebGLM \end{tabular}}} & \\
& French, German, Japanese &   & 3,600 \\
& Korean, Russian, Spanish &   \\
\midrule
\multicolumn{1}{l}{\multirow{3}{*}{ADU-Ambiguity}} & Intonation-based  & \multicolumn{1}{c}{\multirow{3}{*}{\begin{tabular}[c]{@{}c@{}}Phonetics and phonology \\  books \end{tabular}}} &  \\
& Pause-based, Homophone-based & & 1,390 \\
& Repetition-based  & & \\
\bottomrule
\end{tabular}}
\end{table*}



\subsection{Overall}

ADU-Bench is a comprehensive evaluation benchmark designed to assess the open-ended audio dialogue understanding of LALMs in scenarios where LALMs directly respond to user audio inputs. 
ADU-Bench consists of 4 datasets, including ADU-General dataset, ADU-Skill dataset, ADU-Multilingual dataset, and ADU-Ambiguity dataset. 
During data collection, our ADU-Bench contains 20,715 open-ended audio dialogues, comprising over 8,000 real-world recordings alongside synthetic audio samples. The generation details of synthetic audio samples are  in Appendix \ref{app: Generation Details}. The dataset details for ADU-Bench are  in Table \ref{tab:statistics}. Each data point within these datasets is presented as a tuple consisting of (\textit{audio queries},  \textit{textual references}).
The audio queries serve as the input for LALMs, while the textual references function as the expected ground truths. The generation of textual references involves inputting the corresponding textual transcriptions of audio queries into GPT-4 or employing human annotation for ambiguity types. 
A textual format is chosen for the data construction because ADU-Bench focuses on the understanding of audio dialogues instead of generation.

\subsection{Data Construction}

The ADU-General dataset is constructed to evaluate the general dialogue understanding capabilities of LALMs. This dataset comprises 12,000 open-ended audio dialogues, specifically designed to reflect a wide array of inquiries and remarks commonly encountered in life.  It covers  3  scenarios as follows. (1) Helpful questions: These are typically aimed at eliciting useful responses from search engines, such as ``Who won the most gold medals in the Olympics?''. (2) Daily questions: These represent casual questions that arise in real-life conversations, for example, ``What are you doing on this fine day?''. (3) Daily statements: These include everyday remarks, such as ``One today is worth two tomorrows.''. In particular, daily questions and statements are relatively casual without rich contextual information to represent real-world situations.
The construction of this dataset draws from various sources including Alpaca \citep{taori2023stanford}, NQ-Bench \citep{kwiatkowski2019natural}, WebGLM \citep{liu2023webglm}, Slue HVB \citep{shon2022slue}, and Common Voice \citep{ardila2019common}. 
To eliminate queries that do not align with the aforementioned categories, we implement a filtering process combining 
GPT-4 and human inspection.

The ADU-Skill dataset is specifically designed to assess the domain-specific skills of LALMs. This dataset comprises 3,750 audio dialogues and encompasses 12 different domains, including Mathematics, Physics, Chemistry, Biology, Computer Science, Coding, Law, Finance, Common Sense, Writing, Roleplay, and Medicine. To cover these diverse domains, we collect sources for these dialogues from GSM8K \citep{cobbe2021training}, MATH \citep{hendrycks2021measuring}, WizardLM \citep{xu2023wizardlm}, ShareGPT \citep{chiang2023vicuna}, MBPP \citep{austin2021program}, MMLU \citep{hendrycks2020measuring}, HotpotQA \citep{yang2018hotpotqa}, and StrategyQA \citep{geva2021did}. Notably, in certain domains, particularly Mathematics, Physics, and Coding, some queries involve a high volume of Latex formulas or Python code, which can be challenging to comprehend when transformed into audio. Therefore, we employ GPT-4 and human inspection to filter out queries with an excessive number of Latex formulas or Python code. 

The ADU-Multilingual dataset aims to evaluate the multilingual dialogue understanding abilities, covering 9 languages: Arabic, Chinese, English, French, German, Japanese, Korean, Russian, and Spanish. This dataset consists of 3,600 audio dialogues. For generation, we first randomly choose 400 different queries in English from ADU-General dataset. Subsequently, these queries are then translated into the other 8 languages using GPT-4. By including multiple languages, this dataset tests LALMs to understand the audio dialogues in various linguistic contexts. Furthermore, the design of this dataset allows for future expansion, enabling the inclusion of additional languages as needed.


The ADU-Ambiguity dataset is specifically designed to evaluate the robustness of LALMs in addressing ambiguity from different phonetic elements present in audio dialogues. 
It is important to note that ambiguity refers to instances where the textual transcriptions alone, without the accompanying audio or contexts, can lead to confusion. However, when considering the phonetic elements or contextual information provided by the audio, these ambiguities can be resolved, leading to a standard, unambiguous response for humans.
Concretely, this dataset consists of 1,390 audio dialogues, which can be classified into 4 types of ambiguous situations, as described below. (1) Intonation-based ambiguity: In this case, expressing the same sentence with different intonations leads to different interpretations. For instance, ``What a perfect day for the beach.'' can convey different meanings depending on the intonation used. An uplifting intonation indicates that it is indeed a perfect day, while a disappointed intonation signifies that the conditions are far from ideal for a beach day. (2) Pause-based ambiguity:  The placement of pauses can alter the meaning of a sentence. For example, consider the phrase ``professional reviewers and authors.'' Depending on where the pause is placed, it can imply that both the reviewers and authors are smart, or that only the reviewers are smart while the authors are not. The ambiguity arises from the different ways in which pauses can be inserted into the sentence, leading to contrasting interpretations.  (3)  Homophone-based ambiguity: These are sentences containing words that sound almost the same when spoken but have completely different meanings due to variations in word spelling. For example, the words ``weight'' and ``wait'' sound almost the same but convey different meanings.  (4) Repetition-based ambiguity: These sentences contain multiple occurrences of the same word, often leading to confusion. An example of this is, ``I saw a man saw a saw with a saw.''  The construction of the ADU-Ambiguity dataset is achieved manually, drawing upon research studies \citep{mcmahon2002introduction,carr2019english} related to phonetics. To annotate 
textual references, we employ a combination of GPT-4 and manual inspection, ensuring the accuracy and relevance of the references.

\section{Evaluation Method}
\label{sec: Evaluation Method}
Given recent studies \citep{zheng2024judging,yang2024air} have demonstrated that the evaluation with LLMs exhibits better alignment with human preferences, we propose to adopt the advanced LLM, GPT-4, to evaluate the quality of the responses generated by LALMs.
Concretely, LALMs first are queried with audio queries and generate textual responses directly, or convert audio responses into textual format. Subsequently, we present the textual transcriptions of audio queries, textual references (expected ground truths) generated by GPT-4, and textual responses generated by LALMs into the GPT-4 evaluator. Finally, the GPT-4 evaluator assigns an overall score on a scale of 0 to 10 for each data point. 
A higher score indicates the better LALMs' capabilities in handling open-ended audio dialogues. 
The evaluation prompt templates are in Appendix \ref{app: Prompts for Evaluation}. To eliminate the position bias arising from the order of references and responses, we perform a second scoring by swapping their positions and report the average results. Moreover, to avoid bias from GPT-4, we also use LLaMA-3-70B-Instruct and Qwen-2-72B-Instruct for evaluation. 

\section{Results and Analysis}

\subsection{Experimental Settings}
\label{sec: Experimental Settings}
To benchmark the audio dialogue understanding of existing LALMs, we evaluate 16 foundational models with audio understanding capabilities. These models include PandaGPT-7B \citep{su2023pandagpt}, NExT-GPT-7B \citep{wu2023next}, Qwen-Audio-7B \citep{chu2023qwen}, Qwen-Audio-Chat-7B \citep{chu2023qwen}, Mini-Omni-0.5B \citep{xie2024mini}, SpeechGPT-7B \citep{zhang2023speechgpt}, Moshi-7B \citep{defossez2024moshi}, SALMONN-7B \citep{tang2023salmonn}, SALMONN-13B \citep{tang2023salmonn}, BLSP-7B \citep{wang2023blsp}, Step-Audio-Chat-130B \citep{huang2025step}, Whisper-large-v3 \citep{radford2023robust} with LLaMA-2-7B-Chat \citep{touvron2023llama}, with LLaMA-3-8B-Instruct \citep{metallama3},  with LLaMA-3-70B-Instruct  \citep{metallama3}, with GPT-4 (\texttt{gpt-4o-0613}) \citep{achiam2023gpt}, and GPT-4o (\texttt{gpt-4o-audio-preview-2024-12-17}) \citep{openaigpt4o}. 
Unless stated otherwise, the hyperparameters and setups used during the evaluation process remain consistent with those specified in the original papers of the respective models. 
For evaluation, we obtain two evaluation scores by swapping references and responses in the prompts for the GPT-4 evaluator and finally report the average scores for each model in Table \ref{tab:main results}. In addition, to avoid the bias of evaluation only using GPT-4, we apply various open-sourced LLMs for such evaluations, including LLaMA-3-70B-Instruct \citep{metallama3} and Qwen-2-72B-Instruct \citep{chu2023qwen}. In addition, we conduct a direct human evaluation on randomly selected 140 audio dialogues. Each sample is assessed by  three human testers, who rate the generated responses. More details about experimental settings and human evaluation are in Appendix \ref{app: Details of Experimental Settings} and Appendix \ref{app: Human Evaluation Study Details}, respectively. 

\begin{table*}[t]
\caption{The average evaluation scores for audio dialogue understanding under 16 LALMs in our ADU-Bench.
}
\vspace{-1em}
\small
\label{tab:main results}
\centering
\setlength{\tabcolsep}{10.2pt}{
\begin{tabular}{lc|cccc|c|c}
\toprule
\multicolumn{1}{l}{\multirow{2}{*}{Models}} & \multicolumn{1}{l|}{\multirow{2}{*}{Size}} & \multicolumn{4}{c|}{ADU-Bench} & \multicolumn{1}{c|}{\multirow{2}{*}{Average}} & Human \\
&  & General & Skill & Multilingual & Ambiguity & & Evaluation \\
\midrule 
PandaGPT & 7B & 1.02 & 0.98 & 0.98 & 0.50 & 0.87 & - \\
NExT-GPT & 7B & 1.07 & 1.03 & 1.02 & 0.52 & 0.91 & - \\
Qwen-Audio & 7B & 1.32 & 1.08 & 1.07 & 0.61 & 1.02 & - \\
Mini-Omni & 0.5B & 2.31 & 1.96 & 1.55 & 1.67 & 1.87 & - \\
SALMONN & 7B & 2.47 & 2.01 & 1.83 & 1.73 & 2.01 & - \\
Qwen-Audio-Chat & 7B & 2.34 & 2.46 & 1.58 & 1.93 & 2.08 & - \\
SpeechGPT & 7B & 3.99 & 3.56 & 1.42 & 2.25 & 2.81 & - \\
Moshi & 7B & 4.37 & 3.08 & 1.49 & 2.81 & 2.94 & - \\
SALMONN & 13B & 4.07 & 3.12 & 3.25 & 1.86 & 3.08 & - \\
BLSP & 7B & 4.66 & 4.49 & 2.89 & 3.37 & 3.85 & - \\
Step-Audio-Chat & 130B & 6.37 & 7.31 & 2.45 & 4.72 & 5.21 & - \\
\midrule
Whisper+LLaMA-2 & 7B & 6.30 & 6.26 & 4.92 & 4.39 & 5.47 & 6.43 \\
Whisper+LLaMA-3 & 8B & 6.94 & 7.88 & 6.27 & 4.92 & 6.50 & 6.85 \\
Whisper+LLaMA-3 & 70B & 7.26 & 8.03 & 6.12 & 5.13 & 6.64 & 7.46 \\
Whisper+GPT-4 & - & 8.42 & 8.62 & 8.07 & 5.54 & 7.66 & 8.02 \\
\midrule
GPT-4o & - & 8.64 & 8.97 & 8.16 & 6.87 & 8.16 & 8.58 \\
\bottomrule
\end{tabular}}
\end{table*}






\subsection{Main Results}
We report the experimental results for the performance of 16 different LALMs on audio dialogue understanding in Table \ref{tab:main results} and provide a comprehensive analysis of them. Firstly, it can be observed that PandaGPT, NExT-GPT, and Qwen-Audio exhibit the lowest performances, with an average score value of about 1.00. It illustrates that although PandaGPT and NExT-GPT are end-to-end LALMs capable of processing a wide range of modalities, their performances on audio dialogue understanding are relatively lower.
As for Qwen-Audio, a pre-trained base LALM, its weak capabilities in audio dialogue indicate a potential necessity for more specialized training to enhance its understanding of audio dialogues.

Compared to them, Mini-Omni-0.5B, SALMONN-7B and Qwen-Audio-Chat show somewhat superior performance. This can be attributed to the fact that Mini-Omni-0.5B, SALMONN-7B, and Qwen-Audio-Chat have been developed under audio instruction tuning, making them suitable for a variety of audio-oriented scenarios. Moreover, SpeechGPT, Moshi, SALMONN-13B, BLSP, and Step-Audio-Chat have demonstrated even higher proficiency, as reflected in their average scores all about or exceeding 3.00. Among these, BLSP stands out with the highest average score of 3.85 among all LALMs. As SALMONN increases in size from 7B to 13B, its audio dialogue understanding capabilities also show improvement. In addition, both SpeechGPT and BLSP enable audio dialogue with LLMs using speech and exhibit impressive dialogue capabilities. Therefore, it can achieve enhanced performance when using the targeted audio dialogue tuning for end-to-end LALMs.

Furthermore, cascaded LALMs, including LLaMA-2-7B, LLaMA-3-8B, LLaMA-3-70B, and GPT-4 with a Whisper model, obtain higher scores in audio dialogue understanding. Therein, GPT-4 leads the pack with a high score of 7.66.
Following it, LLaMA-3 (including LLaMA-3-8B and LLaMA-3-70B) ranks second, outperforming its predecessor, LLaMA-2. The improved performance of LLaMA-3 to LLaMA-2 highlights the effectiveness of updates  in the LLaMA series.

Notably, for the advanced proprietary LALM, GPT-4o, achieves the highest average score of 8.16, which indicates that it is the best-performing model among the evaluated LALMs.


In addition, experimental results reveal that the GPT-4 evaluator demonstrates a significantly higher correlation with human evaluations, as shown in Table \ref{tab:main results}. 
Furthermore, we also conduct another human evaluation study, detailed in Section \ref{sec: Ablation Study}. These human evaluations verify the alignment between GPT-4 evaluator and human judgments.

\begin{figure*}[t] \centering  
\begin{minipage}{\textwidth}
\includegraphics[width=\textwidth]{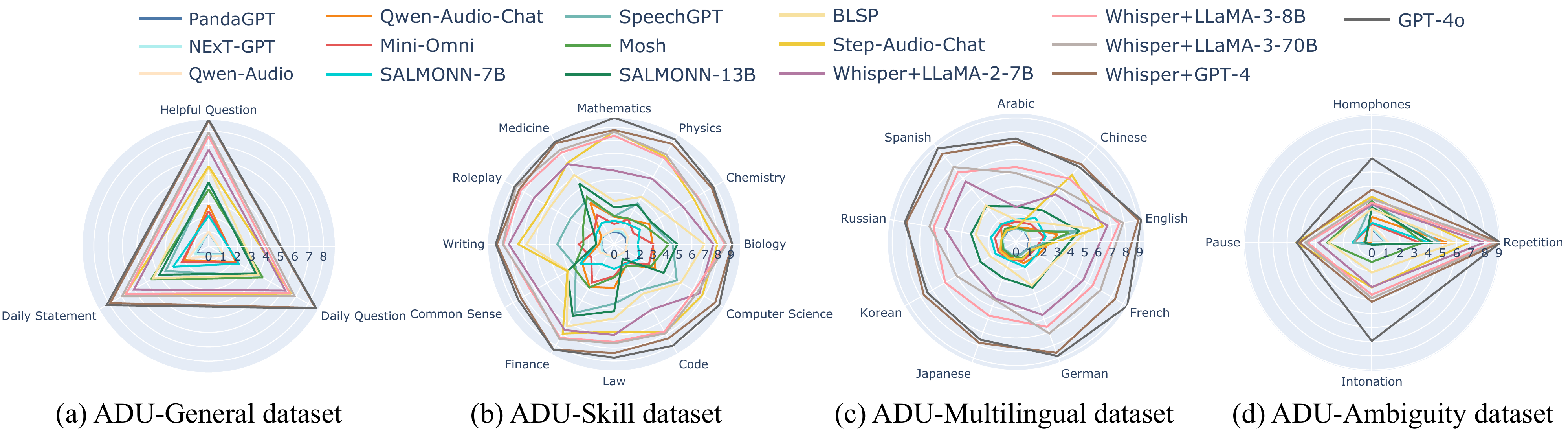}     
\end{minipage}
\vspace{-0.5em}
\caption{The average scores across each domain for 4 datasets within ADU-Bench under 16 LALMs.}
\label{fig:lidar}
\vspace{-1em}
\end{figure*}

\subsection{Results on Each Dataset}

\textbf{Results on ADU-General dataset}.
The ADU-General dataset aims to evaluate the proficiency in general dialogue understanding, with results across 3 scenarios shown in Fig. \ref{fig:lidar}(a). 
Our analysis reveals that LALMs perform better in 
helpful questions compared to daily questions and daily statements. Helpful questions typically seek specific information, whereas daily questions and daily statements represent everyday communication between humans, often characterized by a lack of rich contextual information.
This finding suggests that LALMs are more adept at handling audio dialogues that require precise information retrieval, while their performance in everyday dialogues remains an area for improvement.
In summary, existing open-sourced LALMs understand helpful questions better than daily questions and statements, highlighting the continued development in LALMs to better address everyday human interactions.



\noindent \textbf{Results on ADU-Skill dataset}.
The ADU-Skill dataset is designed to evaluate the skill capabilities of LALMs during audio dialogue and the results across 12 domains are shown in Fig. \ref{fig:lidar}(b). 
Among all these domains, LALMs demonstrate a relative proficiency in handling topics such as Biology, Computer Science, Law, Finance, Writing, and Medicine.  This observation suggests that LALMs possess a certain knowledge foundation in these domains. Meanwhile, these tasks primarily involve language understanding and generation, which align well with the core capabilities of LALMs. 
Moreover, LALMs exhibit weaker performance when dealing with subjects like Mathematics, Physics, Chemistry, and Coding. This can be attributed to the fact that they all involve mathematical symbols and formulas or programming languages so that LALMs struggle to effectively understand these domain-specific challenges they present. 
Additionally, LALMs display limitations in areas related to Common Sense and Roleplay. These domains usually require a deeper understanding of human behavior and LALMs lack the ability to infer implicit meanings or cultural nuances that are crucial for accurately understanding and responding to them.
In summary, existing open-sourced LALMs have knowledge backgrounds in some domains but they face challenges in subjects involving mathematical notations or programming languages, as well as areas requiring a deeper understanding of human behavior.




\noindent \textbf{Results on ADU-Multilingual dataset}.
The ADU-Multilingual dataset aims to evaluate  multilingual capabilities of LALMs during audio dialogues, with results across 9 languages depicted in Fig. \ref{fig:lidar}(c). It can be observed that all LALMs perform best in English due to the massive amount of training data in English. Subsequently, the performance is followed by German, Spanish, French, and Russian. We conjecture that this is because these languages all belong to the Indo-European languages that LALMs can understand to a certain extent. As for other languages, LALMs  exhibit weaker performance which illustrates that they need to be incorporated into the development of LALMs. In conclusion, existing open-sourced LALMs struggle with their multilingual capabilities, highlighting further research to consider various linguistic contexts when developing LALMs.

\begin{figure*}[t] \centering     
\includegraphics[width=\textwidth]{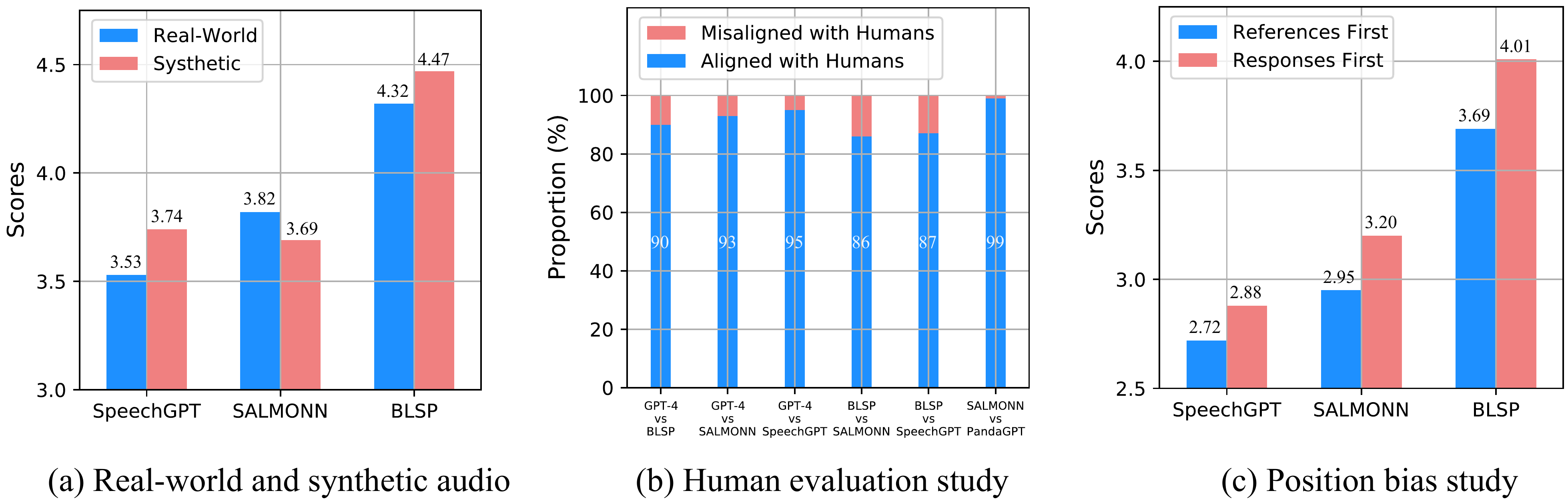} 
\vspace{-2em}
\caption{Ablation study on ADU-Bench. (a) Real-world and synthetic audio can both serve as evaluation sources. (b) GPT-4 evaluator is aligned with human evaluation. (c) Scoring twice is necessary to eliminate the position bias. } 
\label{fig:ablation}
\vspace{-0.5em}
\end{figure*}

\noindent \textbf{Results on ADU-Ambiguity dataset}. 
The ADU-Ambiguity dataset is designed to assess how well LALMs handle 4 types of ambiguity during audio dialogue, including intonation-based, pause-based,  homophone-based, and repetition-based ambiguity, with results in Fig. \ref{fig:lidar}(d). Overall, LALMs exhibit relatively better performance in handling repetition-based ambiguity, while their performance in managing other types of ambiguities is weaker. This observation suggests that LALMs can more effectively resolve ambiguities that do not involve phonetic elements, such as repetition-based ambiguity, which only has multiple words in an audio. However, when it comes to the other three types of ambiguities, including intonation-based, pause-based, and homophone-based, LALMs struggle to handle them effectively. For homophone-based ambiguity, it is difficult for LALMs to distinguish the words that have almost the same pronunciation. For the other two types of ambiguity, LALMs can not perceive the variations in intonations or pause positions, which can lead to expressing different intentions beyond the same literal meaning of sentences. 
When faced with these ambiguities, relatively advanced LALMs like GPT-4o can achieve an average score of 5.22 and 6.05 for pause-based and homophone-based ambiguity. The results show that GPT-4o often generates responses that encompass multiple possible interpretations, which is unable to effectively distinguish between the different meanings based on phonetic elements.
In summary, existing LALMs, including GPT-4o, display limitations in handling the audio dialogue ambiguity in different phonetic elements.




\subsection{Ablation Study}
\label{sec: Ablation Study}

\textbf{Effect of LALMs' size}. We compare the audio dialogue understanding capabilities of SALMONN and LLaMA-3 with a Whisper model with different sizes. As shown in Table \ref{tab:main results}, it indicates a trend of improved average scores with increasing model sizes. However, it is noted that SALMONN-7B outperforms its larger counterpart, SALMONN-13B on Code within ADU-Skill dataset. Similarly, LLaMA-3-8B achieves superior performance than LLaMA-3-70B on Common Sense within ADU-Skill dataset and non-Indo-European languages within ADU-Multilingual dataset. These observations suggest that while a larger model size generally contributes to better overall audio dialogue understanding performance, it can also introduce performance losses in certain domains.

\noindent \textbf{Effect of real-world and synthetic audio}. 
For the audio dialogues difficult to obtain directly, we choose to adopt a synthetic algorithm to generate corresponding audios, as detailed in Appendix \ref{app: Generation Details}. 
To demonstrate that the use of synthetic audio is a feasible approach compared to real-world audio when evaluating LALMs, we randomly sample 1,000 real-world audio dialogues and generate synthetic audio from their transcriptions. The comparison between the real-world audio and the synthetic audio with the same transcriptions is presented in Fig. \ref{fig:ablation}(a). We observe that there is no considerable difference in the performance of LALMs when processing real-world and synthetic audio. In conclusion,  both real-world audio and synthetic audio can effectively serve as evaluation sources for audio dialogue understanding.

\noindent \textbf{Human evaluation study}. 
For evaluation, we choose to adopt GPT-4 as the evaluator.
To evaluate the consistency between the evaluations of GPT-4 and human judgments, we conduct a human evaluation study as follows. Given the challenge of human testers directly assigning a score on a scale of 0 to 10, we adopt a pairwise comparison approach for models, following \citep{touvron2023llama}. Specifically, human testers first listen to the audio queries, then compare two textual responses from two models,
finally indicate their preference as ``A is better'', ``B is better'', or ``Both are equal''. We then convert the GPT-4 scores into the same preference-based rating as the human testers.
Finally, we evaluate the consistency between the two sets of results, as shown in Fig. \ref{fig:ablation}(b).
Our analysis reveals that the pairwise preference consistency achieves a score above 85\%, indicating that GPT-4 evaluation aligns well with human judgments. The details 
are in Appendix \ref{app: Human Evaluation Study Details}. We provide the evaluation results by LLaMA-3-70B-Instruct and Qwen-2-72B-Instruct and the corresponding human evaluation study in Appendix \ref{app: Evaluation by LLaMA-3-70B-Instruct and Qwen-2-72B-Instruct}.

\noindent \textbf{Position bias study}.
To mitigate potential biases from the order of references and responses in the evaluation GPT-4 prompt, we query the GPT-4 evaluator to generate two scores by adjusting their positions. Subsequently, we report the average score for each model. To validate the necessity of scoring twice, we compare the differences between the two scores, presented in Fig. \ref{fig:ablation}(c).
We observe that despite using the same references and responses, the GPT-4 evaluator generates different scores after adjusting the positions. This suggests the existence of a positional bias, particularly when responses are placed before the references.
The observation highlights the importance of conducting a second scoring to address this bias.

\section{Conclusion}
In conclusion, we present ADU-Bench, a comprehensive benchmark designed to evaluate the audio dialogue understanding of LALMs. 
It encompasses 4 benchmark datasets including ADU-General dataset for 3 general scenarios, ADU-Skill dataset for 12 skills, ADU-Multilingual dataset for 9 languages, and ADU-Ambiguity for 4 ambiguity types, providing over 20,000 open-ended audio dialogues for the LALM assessment. 
Our extensive experiments on 16 LALMs reveal that there is still significant room for improvement in their audio dialogue understanding. 
Notably, LALMs face challenges in processing mathematical symbols and formulas, comprehending human behavior like roleplay, understanding multiple languages, and handling audio dialogue ambiguities arising from various phonetic elements. 



\section*{Limitations}
\label{app: Limitations}
The main limitation of this work is that the analysis is on a limited number of LALMs due to the availability of usable code, model weights, and massive experiments. Potential future work includes investigating more diverse LALMs and designing more domains about audio dialogues to make our ADU-Bench up-to-date.

\section*{Ethics Statement}
\label{app: Broader Impacts}

Our ADU-Bench has been carefully curated to ensure that it does not contain any words or content that discriminate against any individual or group. The prompts used in our experiments, as detailed in Appendix \ref{app: Prompts for Evaluation}, have been meticulously reviewed to emphasize that none of them contain any discriminatory language or themes. Moreover, we have taken the necessary precautions to ensure that the prompts used in our work do not negatively impact anyone's safety or well-being. Furthermore, all the codes comply with the MIT License. This commitment to ethical considerations~\citep{deng2024deconstructing} in our research contributes to the responsible development and advancement of LALMs.

\section*{Acknowledgement}
This work is supported in part by the National Natural Science Foundation of China under Grant 62171248 and Shenzhen Science and Technology Program (JCYJ20220818101012025).




\bibliography{custom}
\clearpage

\appendix

\section{Generation Details for Synthetic Datasets}
\label{app: Generation Details}

Our ADU-Bench contains 20,715 open-ended audio dialogues, comprising over 8,000 real-world recordings alongside synthetic audio samples. In this section, we introduce the generation details for the synthetic datasets.

To generate synthetic datasets for ADU-General dataset, ADU-Skill dataset, and ADU-Multilingual dataset, we first adopt GPT-4 and human inspection to obtain the related textual dialogues for each dataset. Then, enclose them in the Speech Synthesis Markup Language (SSML) \citep{taylor1997ssml} by human coding, where SSML is an XML-based markup language specifically designed for speech synthesis applications. Subsequently, execute the program code using Python interpreter with public SSML service \citep{microsoftsslm} provided by Microsoft Azure to convert them into audio dialogues.  Furthermore,  to emulate real-world scenarios, we consider a wide array of variables for synthetic audio. They include 2 genders (male and female), 4 different speakers (2 men and 2 women), 4 emotions (calm, excited, angry, and sad), 3 speech rates (standard and $\pm10\%$), 3 pitch levels (standard and $\pm10\%$), and 3 volume levels (standard and $\pm10\%$). During the generation of each dataset, a combination of these audio generation characteristics is randomly selected to create each audio data, ensuring diversity in the audio dialogues. Therefore, this generation method not only provides a scalable solution for generating synthetic audio datasets but also ensures a rich diversity that closely mirrors real-world audio dialogues.

To construct the ADU-Ambiguity dataset, we first identify four types of ambiguity handling from phonetics and phonology books~\citep{mcmahon2002introduction,carr2019english}. These include ambiguity stemming from intonation, pause positions, homophones, and repetition. Based on the examples and principles outlined in these references, we then manually craft or use GPT-4 to generate many textual data instances representing these ambiguity types.

To convert these textual instances into audio samples, we leverage the Speech Synthesis Markup Language (SSML)~\citep{taylor1997ssml} and use a publicly available SSML service\citep{microsoftsslm}. Specifically:

\begin{itemize}
    \item For intonation-based ambiguity, we use the SSML tags \texttt{<prosody>} to adjust the intonation elements of the audio.
    \item For pause-based ambiguity, we use the SSML tags \texttt{<break>} to insert pauses within the audio.
    \item For homophone-based and repetition-based ambiguity, we are able to directly generate the audio samples without the need for specialized SSML markup.
\end{itemize}

Finally, we conduct a manual validation process to ensure the quality and correctness of the generated audio samples. This involves having human annotators listen to the samples and verify that the intended ambiguity is successfully conveyed through the audio.

\begin{table*}[t]
\caption{Association between human judgment and each dataset in ADU-Bench of GPT-4 evaluation.}
\vspace{-1em}
\small
\label{tab:human judgment gpt}
\centering
\setlength{\tabcolsep}{8.6pt}{
\begin{tabular}{l|cccccc}
\toprule
& \multicolumn{1}{c}{\multirow{2}{*}{\begin{tabular}[c]{@{}c@{}}GPT-4 vs \\ BLSP \end{tabular}}} & \multicolumn{1}{c}{\multirow{2}{*}{\begin{tabular}[c]{@{}c@{}}GPT-4 vs \\ SALMONN \end{tabular}}} & \multicolumn{1}{c}{\multirow{2}{*}{\begin{tabular}[c]{@{}c@{}}GPT-4 vs \\ SpeechGPT \end{tabular}}}  & \multicolumn{1}{c}{\multirow{2}{*}{\begin{tabular}[c]{@{}c@{}}BLSP vs \\ SALMONN \end{tabular}}} & \multicolumn{1}{c}{\multirow{2}{*}{\begin{tabular}[c]{@{}c@{}}BLSP vs \\ SpeechGPT \end{tabular}}} & \multicolumn{1}{c}{\multirow{2}{*}{\begin{tabular}[c]{@{}c@{}}SALMONN vs \\ PandaGPT \end{tabular}}}
 \\ \\
\midrule 
ADU-General & 86.7\% & 80.0\% & 93.3\% & 86.7\% & 93.3\% & 100\% \\
ADU-Skill & 86.7\% & 93.3\% & 93.3\% & 83.3\% &  88.3\% & 100\% \\
ADU-Multilingual & 95.6\% & 95.6\% & 97.8\% & 86.7\% & 86.7\% & 97.8\% \\
ADU-Ambiguity & 90.0\% & 95.0\% & 95.0\% & 90.0\% & 90.0\% & 100\% \\
\midrule 
ADU-Bench & 90.0\% & 92.9\% & 95.0\% & 85.7\% & 87.1\% & 99.3\%
 \\
\bottomrule
\end{tabular}}
\end{table*}

\section{Prompts for Evaluation}
\label{app: Prompts for Evaluation}
The score judgment is based on criteria including helpfulness, relevance, accuracy, and comprehensiveness, comparing the references and generated responses. 
The evaluation prompt for \textit{the first scoring} is as follows.
\begin{tcolorbox}[colback=black!5!white,colframe=black!75!black,title=System Prompt]
You are a helpful and precise assistant for checking the quality of the answer.
\tcbsubtitle{Prompts for Evaluation in ADU-Bench}
Please evaluate the following LALMs' response for the user query and a reference is provided. \\

Query: \textit{Textual Transcriptions} \\
Reference: \textit{Textual References} \\
Response: \textit{Textual Responses} \\
\\
Please rate the helpfulness, relevance, accuracy, and comprehensiveness of the LALMs' response.
Please provide an overall score on a scale of 0 to 10, where a higher score indicates better overall performance.
Do not provide any other output text or explanation. Only provide the score. \\

Output:
\end{tcolorbox}

The evaluation prompt for \textit{the second scoring} is as follows. To eliminate the position bias, we swap the position between responses and references in the evaluation prompt.
\begin{tcolorbox}[colback=black!5!white,colframe=black!75!black,title=System Prompt]
You are a helpful and precise assistant for checking the quality of the answer.
\tcbsubtitle{Prompts for Evaluation in ADU-Bench}
Please evaluate the following LALMs' response for the user query and a reference is provided. \\

Query: \textit{Textual Transcriptions} \\
Response: \textit{Textual Responses} \\
Reference: \textit{Textual References} \\
\\
Please rate the helpfulness, relevance, accuracy, and comprehensiveness of the LALMs' response.
Please provide an overall score on a scale of 0 to 10, where a higher score indicates better overall performance.
Do not provide any other output text or explanation. Only provide the score. \\

Output:
\end{tcolorbox}

\begin{figure*}[t] \centering     
\includegraphics[width=\textwidth]{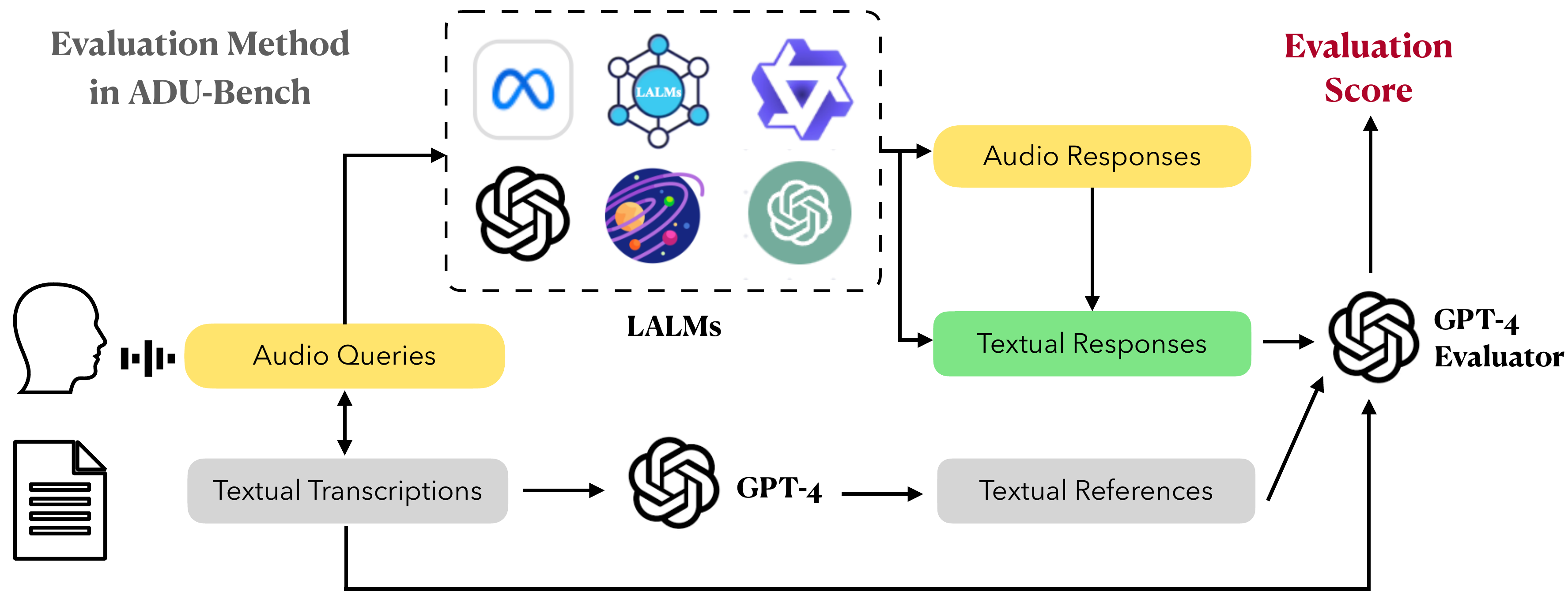} 
\caption{The evaluation method in ADU-Bench. To benchmark open-ended audio dialogue understanding for LALMs, we adopt a GPT-4 evaluator to provide evaluation scores as the metric. We also adopt LLaMA-3-70B-Instruct and Qwen-2-72B-Instruct as the evaluator to provide evaluation scores.} 
\label{fig: evaluation method}
\end{figure*}

The evaluation pipeline is shown in Fig. \ref{fig: evaluation method}. We choose GPT-4 as the default evaluation LLM. We also include LLaMA-3-70B-Instruct and Qwen-2-72B-Instruct to provide the evaluation score. The results are shown in Appendix \ref{app: Evaluation by LLaMA-3-70B-Instruct and Qwen-2-72B-Instruct}.

\section{Details of Experimental Settings}
\label{app: Details of Experimental Settings}
To benchmark the audio dialogue understanding of existing LALMs,  we assess the performance of 16 LALMs across all 4 datasets within ADU-Bench. 
Unless stated otherwise, the hyperparameters and setups used during the evaluation process remain consistent with those specified in the original papers of the respective models. 
For the evaluation, LLaMA-2-7B-Chat, LLaMA-3-8B-Instruct, and LLaMA-3-70B-Instruct are run on 8 NVIDIA A100 40GB GPUs with vLLM library \citep{kwon2023efficient}, while other open-sourced models are run on a single NVIDIA A100 40GB GPU.
By default, our evaluation method employs \texttt{gpt-4-0613} as the GPT-4 evaluator by calling the API. 

\section{Human Evaluation Study Details}
\label{app: Human Evaluation Study Details}

We conduct a direct human evaluation on randomly selected 140 audio dialogues from ADU-Bench. Each sample is assessed by three human testers, who rate the generated responses. Human testers should provide an overall score on a scale of 0 to 10, where a higher score indicates better overall performance. The results are shown in Table \ref{tab:main results}. Besides, we conduct a human evaluation study to evaluate the consistency between the evaluations of GPT-4 and human judgments. We show each pair of samples for ten human testers. The results are demonstrated in Fig. \ref{fig:ablation}(b) and Table \ref{tab:human judgment gpt}. 

For the evaluation datasets, we randomly choose 5 audio queries from each domain in ADU-Bench, and finally obtain 140 audio queries. Since ADU-Multilingual contains multiple languages, it is difficult for human testers to understand each language. Hence, we provide the textual transcriptions and allow them to use the translation tools for evaluation.  we carefully consider the ethical aspects and potential risks associated with the research involving human subjects. The information we collect is only the preference results and does not involve any personal information. 

When selecting participants, there are no requirements for their qualifications, experience, or technical abilities; all participants are adults capable of giving informed consent. We clearly inform the participants of the experiment's content and corresponding compensation before the experiment begins, and we will not cause them any physiological or psychological harm. We randomly select participants within the university campus, informing them of the experiment content, purpose, compensation, and other information. Participants voluntarily decide whether to participate in the experiment after reading the Ethics Informed Consent Form and Ethics Study Information Sheet. The compensation we provide to the participants is 1.5 times the local minimum hourly wage standard.

The instructions given to participants in Table \ref{tab:main results} are as follows:

\textit{Welcome to our human evaluation study! Your participation is crucial in helping us assess the performance of Large Audio-Language Models (LALMs) in audio dialogue understanding.}

\textit{In this study, you will be presented with a total of 140 audio clips, each accompanied by one textual response. For audio in foreign languages, we will provide textual transcriptions and translation tools to assist you. }

\textit{Your task is as follows:}

\textit{1. Listen to the audio queries carefully.}

\textit{2. Based on the criteria of helpfulness, relevance, accuracy, and comprehensiveness, provide an
overall score on a scale of 0 to 10 for the response, where a higher score indicates better overall performance.}

\textit{We appreciate your time and effort in participating in this study. Your valuable insights will significantly contribute to the development and improvement of LALMs, enhancing their ability to understand and respond to audio dialogues effectively. Thank you for your participation!
}

The instructions given to participants in Fig. \ref{fig:ablation}(b) and Table \ref{tab:human judgment gpt} are as follows:

\textit{Welcome to our human evaluation study! Your participation is crucial in helping us assess the performance of Large Audio-Language Models (LALMs) in audio dialogue understanding.}

\textit{In this study, you will be presented with a total of 140 audio clips, each accompanied by two textual responses. For audio in foreign languages, we will provide textual transcriptions and translation tools to assist you. }

\textit{Your task is as follows:}

\textit{1. Listen to the audio queries carefully.}

\textit{2. Compare the two textual responses provided for each audio.}

\textit{3. Based on the criteria of helpfulness, relevance, accuracy, and comprehensiveness, indicate your preference. You can choose from the following options: ``A is better'', ``B is better'', or ``Both are equal''.}

\textit{We appreciate your time and effort in participating in this study. Your valuable insights will significantly contribute to the development and improvement of LALMs, enhancing their ability to understand and respond to audio dialogues effectively. Thank you for your participation!
}

\section{Discussions}
\label{app: Discussions}

\subsection{Real-world and Synthetic Audio in ADU-Bench}
Our ADU-Bench includes both real-world and synthetic audio. As stated in Section \ref{sec: Data Collection and Statistics}, the data collection involves a combination of synthetically generated dialogues and real-world audio samples. Specifically, our ADU-Bench contains over 8000 audio samples from the real world. A reason prevents us from using real-world audio only is the challenges and costs of the collection process. In particular, the collection of professional technical terms or languages can be difficult, as it requires humans who are familiar with them. Without proper familiarity, the use of these terms or languages in audio samples may sound unnatural. To address this issue, we propose a synthetic method for audio generation in Appendix \ref{app: Generation Details}. By leveraging it, we can easily expand a scalable ADU-Bench without incurring substantial expenses. Besides, we have conducted an ablation study to investigate the effects of real-world and synthetic audio on the performance of our benchmark, as detailed in Section \ref{sec: Ablation Study}. It can be observed that there is no significant difference in the performance of LALMs in the areas our ADU-Bench covers. It illustrates that these synthetic audios can also benchmark audio dialogue understanding.

\subsection{Evaluation using Textual Response}
In our evaluation process, we prompt audio queries to obtain audio responses and adopt their textual transcriptions with references to calculate a GPT-4 evaluation score. We have chosen this approach because \textit{our primary focus is on how LALMs comprehend audio dialogue} and formulate appropriate replies. In our ADU-Bench, we emphasize \textit{understanding ability} rather than generation quality of audio dialogues. Directly using audio for evaluation can be challenging, and evaluating generation quality is not within the scope of ADU-Bench. By opting for a textual format, we can concentrate on assessing LALMs' dialogue understanding abilities and their capacity to provide meaningful responses, without introducing the additional complexity of audio generation. Furthermore, our evaluation approach in ADU-Bench aligns with previous work~\citep{yang2024air}.

\subsection{Analysis for Weak Performance of LALMs}
LALMs consist of two main components - audio feature extractors and base LLMs. For textual benchmarks such as GSM8K and MMLU, the base LLMs of LALMs are usually able to achieve effective mathematical and knowledge-based reasoning, which reflects the fundamental reasoning skills of the base LLMs. However, for our ADU-Bench, the LALMs exhibit weak performance and are unable to demonstrate the fundamental reasoning skills of their base LLM components. This observation leads us to conjecture that the poor performance of the LALMs on the ADU-Bench is primarily rooted in their audio comprehension abilities, rather than their core reasoning skills.

\section{Evaluation Results by LLaMA-3-70B-Instruct and Qwen-2-72B-Instruct}
\label{app: Evaluation by LLaMA-3-70B-Instruct and Qwen-2-72B-Instruct}

To avoid the bias of evaluation only using GPT-4, we apply various open-sourced LLMs for such evaluations, including LLaMA-3-70B-Instruct and Qwen-2-72B-Instruct. Our analysis shows that the evaluation scores obtained using these LLMs are mostly consistent with the conclusions drawn from the GPT-4 evaluation. The results are shown in Table \ref{tab:main results by llama3} and Table \ref{tab:main results by qwen2}.

Furthermore, we also include their corresponding human evaluation studies, which can be found in Table \ref{tab:human judgment llama} and Table \ref{tab:human judgment qwen}. All these results indicate that strong LLM evaluations, especially those involving GPT-4, align well with human judgments for audio dialogue understanding. Besides, note that GPT-4 based evaluation is shown to be effective in many areas \citep{zheng2024judging,yang2024air}.


\begin{table*}[t]
\caption{The average evaluation scores under 16 different LALMs on 4 datasets in our ADU-Bench. The evaluation is conducted by LLaMA-3-70B-Instruct.
}
\vspace{-1em}
\small
\label{tab:main results by llama3}
\centering
\setlength{\tabcolsep}{14.5pt}{
\begin{tabular}{lc|cccc|c}
\toprule
\multicolumn{1}{l}{\multirow{2}{*}{Models}} & \multicolumn{1}{l|}{\multirow{2}{*}{Size}} & \multicolumn{4}{c|}{ADU-Bench} & \multicolumn{1}{c}{\multirow{2}{*}{Average}} \\
&  & General & Skill & Multilingual & Ambiguity &  \\
\midrule 
PandaGPT & 7B & 1.00 & 1.00 & 1.00 & 1.00 & 1.00 \\
NExT-GPT & 7B & 1.04 & 1.03 
& 1.00 & 1.00 & 1.02 \\
Qwen-Audio & 7B & 2.00 & 1.00 & 1.42 & 1.00 & 1.36 \\
Mini-Omni & 0.5B & 2.12 & 1.26 & 1.49 & 1.27 & 1.54 \\
SALMONN & 7B & 2.71 & 1.42 & 1.71 & 1.72 & 1.89 \\
Qwen-Audio-Chat & 7B & 1.85 & 3.14 & 2.06 & 1.95 & 2.25 \\
SpeechGPT & 7B & 3.71 & 3.57 & 1.94 & 2.42 & 2.91 \\
Moshi & 7B & 3.96 & 3.15 & 2.04 & 2.64 & 2.95 \\
SALMONN & 13B & 3.71 & 4.23 & 2.92 & 2.05 & 3.23 \\
BLSP & 7B & 4.42 & 3.90 & 2.07 & 2.95 & 3.34  \\
Step-Audio-Chat & 130B & 6.53 & 6.66 & 1.92 & 4.02 & 4.78 \\ 
\midrule
Whisper+LLaMA-2 & 7B & 6.28 & 5.07 & 3.07 & 3.86 & 4.57 \\
Whisper+LLaMA-3 & 8B & 7.57 & 7.00 & 5.00 & 4.75 & 6.08 \\
Whisper+LLaMA-3 & 70B & 7.28 & 7.85 & 6.42 & 5.12 & 6.67 \\
Whisper+GPT-4 & - & 8.57 & 7.92 & 8.50 & 5.46 & 7.61 \\
\midrule
GPT-4o & - & 8.69 & 8.35 & 8.61 & 6.37 & 8.00 \\
\bottomrule
\end{tabular}}
\end{table*}


\begin{table*}[t]
\caption{The average evaluation scores under 16 different LALMs on 4 datasets in our ADU-Bench.  The evaluation is conducted by Qwen-2-72B-Instruct.
}
\vspace{-1em}
\small
\label{tab:main results by qwen2}
\centering
\setlength{\tabcolsep}{14.5pt}{
\begin{tabular}{lc|cccc|c}
\toprule
\multicolumn{1}{l}{\multirow{2}{*}{Models}} & \multicolumn{1}{l|}{\multirow{2}{*}{Size}} & \multicolumn{4}{c|}{ADU-Bench} & \multicolumn{1}{c}{\multirow{2}{*}{Average}} \\
&  & General & Skill & Multilingual & Ambiguity &  \\
\midrule 
PandaGPT & 7B & 1.00 & 1.00 & 1.00 & 1.00 & 1.00 \\
NExT-GPT & 7B & 1.10 & 1.06 & 1.00 & 1.00 & 1.04 \\
Qwen-Audio & 7B & 1.45 & 1.23 & 1.31 & 1.12 & 1.28 \\
Mini-Omni & 0.5B & 1.74 & 1.49 & 1.53 & 1.31 & 1.52 \\
SALMONN & 7B & 2.36 & 1.31 & 2.09 & 1.31 & 1.77 \\
Qwen-Audio-Chat & 7B & 2.57 & 1.74 & 2.45 & 2.85 & 2.40 \\
SpeechGPT & 7B & 4.09 & 4.13 & 2.35 & 2.64 & 3.30 \\
Moshi & 7B & 4.14 & 3.35 & 2.36 & 2.85 & 3.18 \\
SALMONN & 13B & 3.81 & 3.63 & 2.54 & 2.96 & 3.24 \\
BLSP & 7B & 4.18 & 4.54 & 2.48 & 3.84 & 3.76  \\
Step-Audio-Chat & 130B & 6.43 & 5.86 & 2.31 & 3.95 & 4.64 \\ 
\midrule
Whisper+LLaMA-2 & 7B & 6.27 & 5.13 & 3.47 & 3.94 & 4.70 \\
Whisper+LLaMA-3 & 8B & 6.81 & 6.00 & 3.68 & 4.02 & 5.13 \\
Whisper+LLaMA-3 & 70B & 7.18 & 6.63 & 3.86 & 4.36 & 5.51 \\
Whisper+GPT-4 & - & 8.45 & 8.09 & 6.63 & 4.87 & 7.01 \\
\midrule
GPT-4o & - & 8.58 & 8.42 & 6.78 & 5.33 & 7.28 \\
\bottomrule
\end{tabular}}
\end{table*}


\begin{table*}[t]
\caption{Association between human judgment and each dataset in ADU-Bench of of LLaMA-3-70B-Instruct evaluation.}
\vspace{-1em}
\small
\label{tab:human judgment llama}
\centering
\setlength{\tabcolsep}{8.6pt}{
\begin{tabular}{l|cccccc}
\toprule
& \multicolumn{1}{c}{\multirow{2}{*}{\begin{tabular}[c]{@{}c@{}}GPT-4 vs \\ BLSP \end{tabular}}} & \multicolumn{1}{c}{\multirow{2}{*}{\begin{tabular}[c]{@{}c@{}}GPT-4 vs \\ SALMONN \end{tabular}}} & \multicolumn{1}{c}{\multirow{2}{*}{\begin{tabular}[c]{@{}c@{}}GPT-4 vs \\ SpeechGPT \end{tabular}}}  & \multicolumn{1}{c}{\multirow{2}{*}{\begin{tabular}[c]{@{}c@{}}BLSP vs \\ SALMONN \end{tabular}}} & \multicolumn{1}{c}{\multirow{2}{*}{\begin{tabular}[c]{@{}c@{}}BLSP vs \\ SpeechGPT \end{tabular}}} & \multicolumn{1}{c}{\multirow{2}{*}{\begin{tabular}[c]{@{}c@{}}SALMONN vs \\ PandaGPT \end{tabular}}}
 \\ \\
\midrule 
ADU-General & 80.0\% & 86.7\% & 93.3\% & 80.0\% & 86.7\% & 100\% \\
ADU-Skill & 90.0\% & 86.7\% & 93.3\% & 85.0\% & 86.7\% & 98.3\% \\
ADU-Multilingual & 95.6\% & 97.8\% & 97.8\% & 82.2\% & 82.2\% & 100\% \\
ADU-Ambiguity & 90.0\% & 90.0\% & 90.0\% & 86.0\% & 86.0\% & 100\% \\
\midrule 
ADU-Bench & 90.7\% & 90.7\% & 94.3\% & 83.6\% & 85.0\% & 99.3\% \\
\bottomrule
\end{tabular}}
\end{table*}


\begin{table*}[t]
\caption{Association between human judgment and each dataset in ADU-Bench of Qwen-2-72B-Instruct evaluation.}
\vspace{-1em}
\small
\label{tab:human judgment qwen}
\centering
\setlength{\tabcolsep}{8.6pt}{
\begin{tabular}{l|cccccc}
\toprule
& \multicolumn{1}{c}{\multirow{2}{*}{\begin{tabular}[c]{@{}c@{}}GPT-4 vs \\ BLSP \end{tabular}}} & \multicolumn{1}{c}{\multirow{2}{*}{\begin{tabular}[c]{@{}c@{}}GPT-4 vs \\ SALMONN \end{tabular}}} & \multicolumn{1}{c}{\multirow{2}{*}{\begin{tabular}[c]{@{}c@{}}GPT-4 vs \\ SpeechGPT \end{tabular}}}  & \multicolumn{1}{c}{\multirow{2}{*}{\begin{tabular}[c]{@{}c@{}}BLSP vs \\ SALMONN \end{tabular}}} & \multicolumn{1}{c}{\multirow{2}{*}{\begin{tabular}[c]{@{}c@{}}BLSP vs \\ SpeechGPT \end{tabular}}} & \multicolumn{1}{c}{\multirow{2}{*}{\begin{tabular}[c]{@{}c@{}}SALMONN vs \\ PandaGPT \end{tabular}}}
 \\ \\
\midrule 
ADU-General & 80.0\% & 86.7\% & 86.7\% & 80.0\% & 80.0\% & 100\% \\
ADU-Skill & 86.7\% & 90.0\% & 96.0\% & 86.7\% & 80.0\% & 100\% \\
ADU-Multilingual & 93.3\% & 95.6\% & 95.0\% & 82.2\% & 85.0\% & 97.8\%  \\
ADU-Ambiguity & 85.0\% & 90.0\% & 95.0\% & 86.0\% & 85.0\% & 100\% \\
\midrule 
ADU-Bench & 87.9\% & 91.4\% & 95.0\% & 84.3\% & 82.9\% & 99.3\%
 \\
\bottomrule
\end{tabular}}
\end{table*}


\section{Reproducibility Statement}
\label{app: Reproducibility Statement}

We provide the code and data in the project page of our ADU-Bench. 

\section{More Details of ADU-Bench}
\label{app: More Details of ADU-Bench}
The details of ADU-Bench, including the number of each domain within ADU-Bench are in Table \ref{tab: details_adu_bench}.

\begin{table*}[htbp]
\caption{The details of ADU-Bench, including the number of each domain within ADU-Bench.
}
\small
\vspace{-1em}
\label{tab: details_adu_bench}
\centering
\setlength{\tabcolsep}{47.5pt}{
\begin{tabular}{ll|c}
\toprule
\multicolumn{1}{l}{\multirow{1}{*}{Dataset}} & \multicolumn{1}{l|}{\multirow{1}{*}{Domain}} & \multicolumn{1}{c}{Number} \\
\midrule 
\multicolumn{1}{l}{\multirow{3}{*}{ADU-General}} & Helpful Question & 4,000  \\
 & Daily Question & 4,000  \\
 & Daily Statement & 4,000  \\
\midrule 
\multicolumn{1}{l}{\multirow{12}{*}{ADU-Skill}} & Mathematics & 1,000  \\
 & Physics & 210  \\
 & Chemistry & 180  \\
 & Biology & 180  \\
 & Computer Science & 115  \\
 &  Code & 1,000  \\
 & Law & 325  \\
 & Finance & 60  \\
 & Common Sense & 500  \\
 & Writing & 40  \\
 & Roleplay & 20  \\
 & Medicine & 95  \\
\midrule 
\multicolumn{1}{l}{\multirow{9}{*}{ADU-Multilingual}} & Arabic & 400  \\
 & Chinese & 400  \\
 & English  & 400  \\
 & French & 400  \\
 &  German & 400  \\
 & Japanese & 400  \\
 & Korean & 400  \\
 & Russian & 400  \\
 & Spanish & 400  \\
\midrule 
\multicolumn{1}{l}{\multirow{4}{*}{ADU-Ambiguity}} & Intonation-based & 395  \\
 & Pause-based & 250  \\
 & Homophone-based & 490  \\
 & Repetition-based  & 255  \\
\bottomrule
\end{tabular}}
\end{table*}

\section{More Overall Results}
\label{app: More Overall Results}
The overall results of the first and second scoring are shown in Table \ref{tab: app_adu_bench_before} and Table \ref{tab: app_adu_bench_after}.

\begin{table*}[htbp]
\caption{The score for audio dialogue understanding performances under 16 different LALMs on 4 datasets in our proposed ADU-Bench. The textual reference is \textit{before} the textual response in the evaluation prompt for the GPT-4 evaluator.
}
\small
\vspace{-1em}
\label{tab: app_adu_bench_before}
\centering
\setlength{\tabcolsep}{14.7pt}{
\begin{tabular}{lc|cccc|c}
\toprule
\multicolumn{1}{l}{\multirow{2}{*}{Models}} & \multicolumn{1}{l|}{\multirow{2}{*}{Size}} & \multicolumn{4}{c|}{ADU-Bench} & \multicolumn{1}{c}{\multirow{2}{*}{Average}} \\
&  & General & Skill & Multilingual & Ambiguity &  \\
\midrule 
PandaGPT & 7B & 0.98 & 0.97 & 0.97 & 0.49 & 0.85 \\
NExT-GPT & 7B & 1.03 & 0.99 & 0.99 & 0.50 & 0.88 \\
Qwen-Audio & 7B & 1.24 & 0.93 & 0.99 & 0.55 & 0.93 \\
Mini-Omni & 0.5B & 2.20 & 1.87 & 1.49 & 1.51 & 1.77 \\
SALMONN & 7B & 2.35 & 1.92 & 1.71 & 1.69 & 1.92 \\
Qwen-Audio-Chat & 7B & 2.21 & 2.31 & 1.49 & 1.85 & 1.97 \\
SpeechGPT & 7B & 3.91 & 3.40 & 1.39 & 2.18 & 2.72 \\
Moshi & 7B & 4.31 & 3.00 & 1.45 & 2.76 & 2.88 \\
SALMONN & 13B & 3.83 & 3.10 & 3.08 & 1.80 & 2.95 \\
BLSP & 7B & 4.50 & 4.27 & 2.74 & 3.25 & 3.69 \\
Step-Audio-Chat & 130B & 6.30 & 7.23 & 2.39 & 4.66 & 5.15 \\ 
\midrule
Whisper+LLaMA-2 & 7B & 6.07 & 6.20 & 4.82 & 4.30 & 5.35 \\
Whisper+LLaMA-3 & 8B & 6.66 & 7.80 & 6.21 & 4.79 & 6.37 \\
Whisper+LLaMA-3 & 70B & 6.82 & 7.97 & 6.09 & 4.97 & 6.46 \\
Whisper+GPT-4 & - & 8.33 & 8.54 & 8.04 & 5.43 & 7.59 \\
\midrule
GPT-4o & - & 8.54 & 8.84 & 8.07 & 6.79 & 8.06 \\
\bottomrule
\end{tabular}}
\end{table*}

\begin{table*}[htbp]
\caption{The score for audio dialogue understanding performances under 16 different LALMs on 4 datasets in our proposed ADU-Bench. The textual reference is \textit{after} the textual response in the evaluation prompt for the GPT-4 evaluator.
}
\small
\vspace{-1em}
\label{tab: app_adu_bench_after}
\centering
\setlength{\tabcolsep}{14.7pt}{
\begin{tabular}{lc|cccc|c}
\toprule
\multicolumn{1}{l}{\multirow{2}{*}{Models}} & \multicolumn{1}{l|}{\multirow{2}{*}{Size}} & \multicolumn{4}{c|}{ADU-Bench} & \multicolumn{1}{c}{\multirow{2}{*}{Average}} \\
&  & General & Skill & Multilingual & Ambiguity &  \\
\midrule 
PandaGPT & 7B & 1.06 & 0.98 & 0.98 & 0.50 & 0.88 \\
NExT-GPT & 7B & 1.11 & 1.07 & 1.04 & 0.53 & 0.94 \\
Qwen-Audio & 7B & 1.40 & 1.23 & 1.14 & 0.67 & 1.11\\
Mini-Omni & 0.5B & 2.42 & 2.06 & 1.61 & 1.84 & 1.98 \\
SALMONN & 7B & 2.59 & 2.09 & 1.94 & 1.77 & 2.10 \\
Qwen-Audio-Chat & 7B & 2.47 & 2.60 & 1.66 & 2.00 & 2.19 \\
SpeechGPT & 7B & 4.06 & 3.71 & 1.44 & 2.32 & 2.88  \\
Moshi & 7B & 4.43 & 3.16 & 1.52 & 2.86 & 2.99 \\
SALMONN & 13B & 4.31 & 3.14 & 3.42 & 1.91 & 3.20 \\
BLSP & 7B & 4.82 & 4.70 & 3.04 & 3.48 & 4.01 \\
Step-Audio-Chat & 130B & 6.44 & 7.38 & 2.51 & 4.77 & 5.27 \\
\midrule
Whisper+LLaMA-2 & 7B & 6.53 & 6.32 & 5.02 & 4.48 & 5.59 \\
Whisper+LLaMA-3 & 8B & 7.21 & 7.96 & 6.32 & 5.04 & 6.63 \\
Whisper+LLaMA-3 & 70B & 7.70 & 8.09 & 6.14 & 5.29 & 6.81 \\
Whisper+GPT-4 & - & 8.51 & 8.70 & 8.09 & 5.64 & 7.74 \\
\midrule
GPT-4o & - & 8.74 & 9.10 & 8.24 & 6.94 & 8.25 \\
\bottomrule
\end{tabular}}
\end{table*}

\section{More Results on Each Dataset}
\label{app: More Results on Each Dataset}
The results on each dataset of the first and second scoring are shown in Table \ref{tab: app_adu_general_before}, Table \ref{tab: app_adu_general_after}, Table \ref{tab: app_adu_skill_before_part1}, Table \ref{tab: app_adu_skill_before_part2}, Table \ref{tab: app_adu_skill_after_part1}, Table \ref{tab: app_adu_skill_after_part2}, Table \ref{tab: app_adu_multilingual_before_part1}, Table \ref{tab: app_adu_multilingual_before_part2}, Table \ref{tab: app_adu_multilingual_after_part1}, Table \ref{tab: app_adu_multilingual_after_part2}, Table \ref{tab: app_adu_ambiguity_before}, and Table \ref{tab: app_adu_ambiguity_after}.

\begin{table*}[htbp]
\caption{The score for audio dialogue understanding performances under 16 different LALMs on ADU-General dataset. The textual reference is \textit{before} the textual response in the evaluation prompt for the GPT-4 evaluator.
}
\small
\vspace{-1em}
\label{tab: app_adu_general_before}
\centering
\setlength{\tabcolsep}{18.7pt}{
\begin{tabular}{lc|ccc}
\toprule
\multicolumn{1}{l}{\multirow{2}{*}{Models}} & \multicolumn{1}{l|}{\multirow{2}{*}{Size}} & \multicolumn{3}{c}{ADU-General} \\
&  & Helpful Question & Daily Question & Daily Statement  \\
\midrule 
PandaGPT & 7B & 0.99 & 1.00 & 0.96  \\
NExT-GPT & 7B & 1.00 & 1.10 & 1.00 \\
Qwen-Audio & 7B & 0.90 & 1.23 & 1.58 \\
Mini-Omni & 0.5B & 2.34 & 2.24 & 2.02 \\
SALMONN & 7B & 2.05 & 2.34 & 2.66  \\
Qwen-Audio-Chat & 7B & 2.77 & 2.00 & 1.86  \\
SpeechGPT & 7B & 4.37 & 4.09 & 3.28 \\
Moshi & 7B & 3.96 & 4.35 & 4.62  \\
SALMONN & 13B & 4.19 & 3.59 & 3.70 \\
BLSP & 7B & 5.33 & 3.91 & 4.27  \\
Step-Audio-Chat & 130B & 5.56 & 6.61 & 6.73 \\
\midrule
Whisper+LLaMA-2 & 7B & 6.69 & 5.88 & 5.64 \\
Whisper+LLaMA-3 & 8B & 7.65 & 6.12 & 6.22 \\
Whisper+LLaMA-3 & 70B & 7.71 & 6.34 & 6.42  \\
Whisper+GPT-4 & - & 8.63 & 8.51 & 7.84 \\
\midrule
GPT-4o & - & 8.76 & 8.65 & 8.20 \\
\bottomrule
\end{tabular}}
\end{table*}

\begin{table*}[htbp]
\caption{The score for audio dialogue understanding performances under 16 different LALMs on ADU-General dataset. The textual reference is \textit{after} the textual response in the evaluation prompt for the GPT-4 evaluator.
}
\small
\vspace{-1em}
\label{tab: app_adu_general_after}
\centering
\setlength{\tabcolsep}{18.8pt}{
\begin{tabular}{lc|ccc}
\toprule
\multicolumn{1}{l}{\multirow{2}{*}{Models}} & \multicolumn{1}{l|}{\multirow{2}{*}{Size}} & \multicolumn{3}{c}{ADU-General}  \\
&  & Helpful Question & Daily Question & Daily Statement \\
\midrule 
PandaGPT & 7B & 0.99 & 1.17 & 1.03  \\
NExT-GPT & 7B & 0.98 & 1.15 & 1.21 \\
Qwen-Audio & 7B & 1.15 & 1.34 & 1.72 \\
Mini-Omni & 0.5B & 2.56 & 2.43 & 2.26 \\
SALMONN & 7B & 2.20 & 2.51 & 3.06  \\
Qwen-Audio-Chat & 7B &  2.96 & 2.35 & 2.10 \\
SpeechGPT & 7B & 4.39 & 4.12 & 3.66 \\
Moshi & 7B & 4.12 & 4.41 & 4.75\\
SALMONN & 13B & 4.70 & 4.02 & 4.22 \\
BLSP & 7B & 5.64 & 4.14 & 4.68  \\
Step-Audio-Chat & 130B & 5.68 & 6.82 & 6.84 \\
\midrule
Whisper+LLaMA-2 & 7B & 6.75 & 6.46 & 6.38 \\
Whisper+LLaMA-3 & 8B & 7.67 & 6.88 & 7.08 \\
Whisper+LLaMA-3 & 70B & 8.12 & 7.45 & 7.52  \\
Whisper+GPT-4 & - & 8.86 & 8.67 & 8.00 \\
\midrule
GPT-4o & - & 8.92 & 8.74 & 8.55 \\
\bottomrule
\end{tabular}}
\end{table*}

\begin{table*}[htbp]
\caption{The score for audio dialogue understanding performances under 16 different LALMs on ADU-Skill dataset. The textual reference is \textit{before} the textual response in the evaluation prompt for the GPT-4 evaluator.
}
\small
\vspace{-1em}
\label{tab: app_adu_skill_before_part1}
\centering
\setlength{\tabcolsep}{8.5pt}{
\begin{tabular}{lc|cccccc}
\toprule
\multicolumn{1}{l}{\multirow{2}{*}{Models}} & \multicolumn{1}{l|}{\multirow{2}{*}{Size}} & \multicolumn{6}{c}{ADU-Skill (Part I)} \\
&  & Mathematics & Physics & Chemistry & Biology & Computer Science & Code  \\
\midrule 
PandaGPT & 7B & 0.98 & 1.00 & 0.99 & 1.00 & 0.97 & 0.90  \\
NExT-GPT & 7B & 0.99 & 1.02 & 1.14 & 0.98 & 0.99 & 0.96 \\
Qwen-Audio & 7B & 1.03 & 1.19 & 1.04 & 0.86 & 0.89 & 0.84 \\
Mini-Omni & 0.5B & 1.48 & 2.06 & 1.63 & 2.92 & 2.97 & 1.55 \\
SALMONN & 7B & 1.78 & 1.73 & 2.26 & 1.87 & 2.09 & 1.66  \\
Qwen-Audio-Chat & 7B & 1.99 & 2.06 & 2.96 & 2.79 & 3.62 & 1.74  \\
SpeechGPT & 7B & 1.99 & 3.41 & 3.14 & 4.52 & 5.33 & 3.94 \\
Moshi & 7B & 2.17 & 2.28 & 2.77 & 4.19 & 3.33 & 1.94 \\
SALMONN & 13B & 3.15 & 3.24 & 3.09 & 4.76 & 4.31 & 1.31 \\
BLSP & 7B & 2.99 & 3.94 & 4.39 & 6.91 & 5.76 & 4.31  \\
Step-Audio-Chat & 130B & 8.70 & 7.81 & 7.06 & 7.96 & 7.85 & 7.85 \\
\midrule
Whisper+LLaMA-2 & 7B & 5.65 & 5.59 & 5.86 & 7.59 & 7.41 & 5.78 \\
Whisper+LLaMA-3 & 8B & 8.21 & 7.65 & 7.35 & 8.58 & 7.12 & 7.73 \\
Whisper+LLaMA-3 & 70B & 8.63 & 7.93 & 7.38 & 8.62 & 7.21 & 7.84  \\
Whisper+GPT-4 & - & 8.72 & 8.93 & 8.66 & 9.00 & 8.96 & 8.34 \\
\midrule
GPT-4o & - & 9.53 & 9.06 & 8.67 & 8.98 & 9.21 & 8.84 \\
\bottomrule
\end{tabular}}
\end{table*}

\begin{table*}[htbp]
\caption{The score for audio dialogue understanding performances under 16 different LALMs on ADU-Skill dataset. The textual reference is \textit{before} the textual response in the evaluation prompt for the GPT-4 evaluator.
}
\small
\vspace{-1em}
\label{tab: app_adu_skill_before_part2}
\centering
\setlength{\tabcolsep}{10.5pt}{
\begin{tabular}{lc|cccccc}
\toprule
\multicolumn{1}{l}{\multirow{2}{*}{Models}} & \multicolumn{1}{l|}{\multirow{2}{*}{Size}} & \multicolumn{6}{c}{ADU-Skill (Part II)} \\
&  & Law & Finance & Common Sense & Writing &	Roleplay & Medicine  \\
\midrule 
PandaGPT & 7B & 1.01 & 0.98 & 0.99 & 0.97 & 1.00 & 1.00  \\
NExT-GPT & 7B & 0.98 & 1.12 & 1.00 & 0.99 & 1.05 & 0.99 \\
Qwen-Audio & 7B & 0.88 & 0.77 & 0.84 & 1.36 & 1.10 & 0.83 \\
Mini-Omni & 0.5B & 2.39 & 3.31 & 1.96 & 2.64 & 1.72 & 2.52 \\
SALMONN & 7B & 1.84 & 1.82 & 2.81 & 1.39 & 1.56 & 1.85  \\
Qwen-Audio-Chat & 7B & 3.20 & 3.65 & 2.65 & 1.19 & 0.80 & 3.49  \\
SpeechGPT & 7B & 4.40 & 6.08 & 3.22 & 4.50 & 3.52 & 3.93 \\
Moshi & 7B & 2.55 & 3.85 & 3.60 & 2.37 & 2.76 & 4.21 \\
SALMONN & 13B & 4.93 & 6.09 & 3.90 & 1.44 & 1.65 & 5.23 \\
BLSP & 7B & 5.52 & 7.10 & 3.87 & 6.63 & 5.07 & 5.97  \\
Step-Audio-Chat & 130B & 6.74 & 7.95 & 4.16 & 7.41 & 6.12 & 7.23 \\
\midrule
Whisper+LLaMA-2 & 7B & 6.87 & 7.60 & 6.77 & 8.20 & 6.68 & 7.05 \\
Whisper+LLaMA-3 & 8B & 7.44 & 8.35 & 7.26 & 8.42 & 8.24 & 8.10 \\
Whisper+LLaMA-3 & 70B & 7.59 & 8.46 & 7.16 & 8.55 & 8.64 & 8.26  \\
Whisper+GPT-4 & - & 8.25 & 9.38 & 8.12 & 8.92 & 8.12 & 8.93 \\
\midrule
GPT-4o & - & 8.41 & 9.25 & 8.32 & 8.71 & 8.14 & 8.98 \\
\bottomrule
\end{tabular}}
\end{table*}

\begin{table*}[htbp]
\caption{The score for audio dialogue understanding performances under 16 different LALMs on ADU-Skill dataset. The textual reference is \textit{after} the textual response in the evaluation prompt for the GPT-4 evaluator.
}
\small
\vspace{-1em}
\label{tab: app_adu_skill_after_part1}
\centering
\setlength{\tabcolsep}{8.5pt}{
\begin{tabular}{lc|cccccc}
\toprule
\multicolumn{1}{l}{\multirow{2}{*}{Models}} & \multicolumn{1}{l|}{\multirow{2}{*}{Size}} & \multicolumn{6}{c}{ADU-Skill (Part I)} \\
&  & Mathematics & Physics & Chemistry & Biology & Computer Science & Code  \\
\midrule 
PandaGPT & 7B & 0.98 & 1.00 & 1.00 & 0.98 & 1.01 & 0.95  \\
NExT-GPT & 7B & 1.12 & 1.20 & 1.15  & 1.10 & 0.99 & 0.98 \\
Qwen-Audio & 7B & 1.26 & 1.57 & 1.37 & 1.11 & 1.18 & 0.97 \\
Mini-Omni & 0.5B & 1.65 & 2.33 & 1.79 & 3.14 & 3.22 & 1.77 \\
SALMONN & 7B & 1.85 & 1.97 & 2.30 & 1.81 & 2.41 & 1.84  \\
Qwen-Audio-Chat & 7B & 2.25 & 2.34 & 3.29 & 3.16 & 3.69 & 2.00  \\
SpeechGPT & 7B & 2.31 & 3.87 & 3.40 & 4.52 & 5.82 & 4.22 \\
Moshi & 7B & 2.44 & 2.46 & 2.85 & 4.22 & 3.52 & 2.27 \\
SALMONN & 13B & 2.54 & 3.81 & 3.61 & 4.97 & 4.51 & 1.30 \\
BLSP & 7B & 3.68 & 4.50 & 4.81 & 7.00 & 6.12 & 4.51  \\
Step-Audio-Chat & 130B & 8.73 & 7.92 & 7.36 & 8.15 & 8.02 & 7.74 \\
\midrule
Whisper+LLaMA-2 & 7B & 5.71 & 6.08 & 6.17 & 7.70 & 7.77 & 5.82 \\
Whisper+LLaMA-3 & 8B & 8.53 & 7.71 & 7.47 & 8.50 & 7.16 & 7.82 \\
Whisper+LLaMA-3 & 70B & 8.70 & 8.07 & 7.29 & 8.69 & 7.62 & 8.01  \\
Whisper+GPT-4 & - & 8.90 & 8.94 & 8.72 & 9.21 & 9.03 & 8.41 \\
\midrule
GPT-4o & - & 9.76 & 9.35 & 8.84 & 9.12 & 9.36 & 9.03 \\
\bottomrule
\end{tabular}}
\end{table*}

\begin{table*}[htbp]
\caption{The score for audio dialogue understanding performances under 16 different LALMs on ADU-Skill dataset. The textual reference is \textit{after} the textual response in the evaluation prompt for the GPT-4 evaluator.
}
\small
\vspace{-1em}
\label{tab: app_adu_skill_after_part2}
\centering
\setlength{\tabcolsep}{10.5pt}{
\begin{tabular}{lc|cccccc}
\toprule
\multicolumn{1}{l}{\multirow{2}{*}{Models}} & \multicolumn{1}{l|}{\multirow{2}{*}{Size}} & \multicolumn{6}{c}{ADU-Skill (Part II)} \\
&  & Law & Finance & Common Sense & Writing &	Roleplay & Medicine  \\
\midrule 
PandaGPT & 7B & 1.02 & 1.00 & 0.99 & 1.00 & 1.05 & 1.00 \\
NExT-GPT & 7B & 1.00 & 1.03 & 1.12 & 0.99 & 1.12 & 1.00 \\
Qwen-Audio & 7B & 1.08 & 1.13 & 1.65 & 1.40 & 1.27 & 1.16 \\
Mini-Omni & 0.5B & 2.52 & 3.53 & 2.11 & 2.82 & 1.93 & 2.68 \\
SALMONN & 7B & 1.96 & 1.77 & 3.27 & 1.40 & 1.65 & 1.96  \\
Qwen-Audio-Chat & 7B & 3.51 & 3.94 & 3.08 & 1.14 & 1.50 & 3.87  \\
SpeechGPT & 7B & 4.78 & 6.14 & 3.61 & 4.28 & 4.29 & 4.21 \\
Moshi & 7B & 2.77 & 4.02 & 3.85 & 2.53 & 2.84 & 4.13 \\
SALMONN & 13B & 5.39 & 6.67 & 4.44 & 1.32 & 2.00 & 5.58 \\
BLSP & 7B & 5.92 & 7.53 & 4.31 & 6.89 & 6.35 & 6.37  \\
Step-Audio-Chat & 130B & 6.85 & 8.16 & 4.27 & 7.63 & 6.33 & 7.42 \\
\midrule
Whisper+LLaMA-2 & 7B & 7.44 & 8.35 & 7.26 & 8.42 & 8.24 & 8.10 \\
Whisper+LLaMA-3 & 8B & 7.61 & 8.33 & 7.42 & 8.66 & 8.40 & 8.22 \\
Whisper+LLaMA-3 & 70B & 7.68 & 8.42 & 7.29 & 8.64 & 8.89 & 8.51  \\
Whisper+GPT-4 & - & 8.54 & 9.36 & 8.46 & 9.16 & 8.78 & 9.07 \\
\midrule
GPT-4o & - & 8.74 & 9.38 & 8.54 & 9.14 & 8.87 & 9.12 \\
\bottomrule
\end{tabular}}
\end{table*}

\begin{table*}[htbp]
\caption{The score for audio dialogue understanding performances under 16 different LALMs on ADU-Multilingual dataset. The textual reference is \textit{before} the textual response in the evaluation prompt for the GPT-4 evaluator.
}
\small
\vspace{-1em}
\label{tab: app_adu_multilingual_before_part1}
\centering
\setlength{\tabcolsep}{16.3pt}{
\begin{tabular}{lc|ccccc}
\toprule
\multicolumn{1}{l}{\multirow{2}{*}{Models}} & \multicolumn{1}{l|}{\multirow{2}{*}{Size}} & \multicolumn{5}{c}{ADU-Multilingual (Part I)} \\
&  & Arabic & Chinese & English & French & German  \\
\midrule 
PandaGPT & 7B & 0.98 & 0.98 & 0.96 & 0.98 & 0.97  \\
NExT-GPT & 7B & 0.99 & 0.99 & 1.00 & 1.00 & 0.99 \\
Qwen-Audio & 7B & 0.95 & 1.08 & 0.93 & 1.02 & 0.94 \\
Mini-Omni & 0.5B & 1.37 & 1.57 & 1.99 & 1.38 & 1.40 \\
SALMONN & 7B & 1.47 & 2.14 & 2.11 & 1.67 & 1.85  \\
Qwen-Audio-Chat & 7B & 1.00 & 1.18 & 2.95 & 1.73 & 1.54  \\
SpeechGPT & 7B & 0.98 & 1.04 & 4.48 & 1.01 
& 1.00 \\
Moshi & 7B & 1.07 & 1.08 & 3.94 & 1.30 & 1.27 \\
SALMONN & 13B & 2.38 & 2.88 & 4.48 & 2.90 & 3.30 \\
BLSP & 7B & 1.51 & 1.81 & 5.28 & 2.94 & 3.20  \\
Step-Audio-Chat & 130B & 1.00 & 6.34 & 6.43 & 1.24 & 1.36 \\
\midrule
Whisper+LLaMA-2 & 7B & 2.36	& 4.36 & 6.68 & 5.60 & 5.62 \\
Whisper+LLaMA-3 & 8B & 5.33 & 5.97 & 7.56 & 6.36 & 6.50 \\
Whisper+LLaMA-3 & 70B & 5.02 & 5.02 & 7.89 & 7.02 & 7.08  \\
Whisper+GPT-4 & - & 7.26 & 7.34 & 8.99 & 8.32 & 8.60 \\
\midrule
GPT-4o & - & 7.40 & 6.80 & 9.09 & 9.25 & 8.69 \\
\bottomrule
\end{tabular}}
\end{table*}

\begin{table*}[htbp]
\caption{The score for audio dialogue understanding performances under 16 different LALMs on ADU-Multilingual dataset. The textual reference is \textit{before} the textual response in the evaluation prompt for the GPT-4 evaluator.
}
\small
\vspace{-1em}
\label{tab: app_adu_multilingual_before_part2}
\centering
\setlength{\tabcolsep}{20.7pt}{
\begin{tabular}{lc|cccc}
\toprule
\multicolumn{1}{l}{\multirow{2}{*}{Models}} & \multicolumn{1}{l|}{\multirow{2}{*}{Size}} & \multicolumn{4}{c}{ADU-Multilingual (Part II)} \\
&  & Japanese & Korean & Russian & Spanish  \\
\midrule 
PandaGPT & 7B & 0.98 & 0.98 & 0.98 & 0.96  \\
NExT-GPT & 7B & 0.98 & 0.97 & 0.98 & 0.98 \\
Qwen-Audio & 7B & 0.91 & 1.10 & 0.98 & 0.99 \\
Mini-Omni & 0.5B & 1.36 & 1.41 & 1.49 & 1.47 \\
SALMONN & 7B & 1.37 & 1.59 & 1.70 & 1.52  \\
Qwen-Audio-Chat & 7B & 1.08 & 1.16 & 1.01 & 1.75  \\
SpeechGPT & 7B & 1.00 & 1.01 & 1.03 & 1.00 \\
Moshi & 7B & 1.08 & 1.07 & 1.03 & 1.24  \\
SALMONN & 13B & 2.62 & 2.87 & 3.12 & 3.16 \\
BLSP & 7B & 1.86 & 2.00 & 2.80 & 3.27  \\
Step-Audio-Chat & 130B & 1.36 & 1.33 & 1.07 & 1.07  \\
\midrule
Whisper+LLaMA-2 & 7B & 4.25 & 3.73 & 5.20 & 5.60 \\
Whisper+LLaMA-3 & 8B & 5.65 & 5.97 & 6.04 & 6.53 \\
Whisper+LLaMA-3 & 70B & 4.44 & 4.96 & 6.34 & 7.05 \\
Whisper+GPT-4 & - & 7.81 & 7.68 & 8.07 & 8.31 \\
\midrule
GPT-4o & - & 7.34 & 7.28 & 8.01 & 8.73 \\
\bottomrule
\end{tabular}}
\end{table*}

\begin{table*}[htbp]
\caption{The score for audio dialogue understanding performances under 16 different LALMs on ADU-Multilingual dataset. The textual reference is \textit{after} the textual response in the evaluation prompt for the GPT-4 evaluator.
}
\small
\vspace{-1em}
\label{tab: app_adu_multilingual_after_part1}
\centering
\setlength{\tabcolsep}{16.3pt}{
\begin{tabular}{lc|ccccc}
\toprule
\multicolumn{1}{l}{\multirow{2}{*}{Models}} & \multicolumn{1}{l|}{\multirow{2}{*}{Size}} & \multicolumn{5}{c}{ADU-Multilingual (Part I)} \\
&  & Arabic & Chinese & English & French & German  \\
\midrule 
PandaGPT & 7B & 0.99 & 0.97 & 0.98 & 0.98 & 0.98  \\
NExT-GPT & 7B & 0.98 & 1.00 & 1.15 & 1.12 & 1.01 \\
Qwen-Audio & 7B & 1.09 & 1.29 & 1.12 & 1.08 & 1.13 \\
Mini-Omni & 0.5B & 1.55 & 1.76 & 2.12 & 1.47 & 1.52 \\
SALMONN & 7B & 1.76 & 2.32 & 2.27 & 1.92 & 2.05  \\
Qwen-Audio-Chat & 7B & 1.07 & 1.41 & 3.23 & 2.04 & 1.77  \\
SpeechGPT & 7B & 1.00 & 1.10 & 4.68 & 1.04 & 1.03 \\
Moshi & 7B & 1.16 & 1.23 & 4.21 & 1.48 & 1.43 \\
SALMONN & 13B & 2.76 & 3.08 & 4.83 & 3.25 & 3.81 \\
BLSP & 7B & 1.67 & 1.99 & 5.74 & 3.26 & 3.60  \\
Step-Audio-Chat & 130B & 1.21 & 6.56 & 6.57 & 1.45 & 1.42 \\
\midrule
Whisper+LLaMA-2 & 7B & 2.68 & 4.61 & 6.82 & 5.67 & 5.71 \\
Whisper+LLaMA-3 & 8B & 5.53 & 6.03 & 7.69 & 6.50 & 6.68 \\
Whisper+LLaMA-3 & 70B & 4.98 & 5.06 & 7.93 & 7.15 & 7.11  \\
Whisper+GPT-4 & - & 7.28 & 7.39 & 9.10 & 8.33 & 8.60 \\
\midrule
GPT-4o & - & 7.51 & 7.12 & 9.24 & 9.35 & 8.84 \\
\bottomrule
\end{tabular}}
\end{table*}

\begin{table*}[htbp]
\caption{The score for audio dialogue understanding performances under 16 different LALMs on ADU-Multilingual dataset. The textual reference is \textit{after} the textual response in the evaluation prompt for the GPT-4 evaluator.
}
\small
\vspace{-1em}
\label{tab: app_adu_multilingual_after_part2}
\centering
\setlength{\tabcolsep}{20.5pt}{
\begin{tabular}{lc|cccc}
\toprule
\multicolumn{1}{l}{\multirow{2}{*}{Models}} & \multicolumn{1}{l|}{\multirow{2}{*}{Size}} & \multicolumn{4}{c}{ADU-Multilingual (Part II)} \\
&  & Japanese & Korean & Russian & Spanish  \\
\midrule 
PandaGPT & 7B & 0.98 & 0.98 & 0.99 & 0.98  \\
NExT-GPT & 7B & 0.99 & 0.98 & 0.99 & 1.10 \\
Qwen-Audio & 7B & 1.10 & 1.33 & 1.06 & 1.08 \\
Mini-Omni & 0.5B & 1.43 & 1.53 & 1.60 & 1.53 \\
SALMONN & 7B & 1.48 & 1.87 & 1.99 & 1.80  \\
Qwen-Audio-Chat & 7B & 1.31 & 1.37 & 1.04 & 1.69  \\
SpeechGPT & 7B & 1.02 & 1.05 & 1.05 & 1.02 \\
Moshi & 7B &1.24 & 1.21 & 1.16 & 1.42 \\
SALMONN & 13B & 2.96 & 3.06 & 3.52 & 3.58 \\
BLSP & 7B &  2.07 & 2.29 & 3.16 & 3.61 \\
Step-Audio-Chat & 130B & 1.47 & 1.52 & 1.21 & 1.18 \\
\midrule
Whisper+LLaMA-2 & 7B & 5.65 & 5.97 & 6.04 & 6.53 \\
Whisper+LLaMA-3 & 8B & 5.80 & 5.92 & 6.12 & 6.63 \\
Whisper+LLaMA-3 & 70B & 4.54 & 4.98 & 6.44 & 7.11  \\
Whisper+GPT-4 & - & 7.84 & 7.81 & 8.13 & 8.37 \\
\midrule
GPT-4o & - & 7.58 & 7.43 & 8.21 & 8.85 \\
\bottomrule
\end{tabular}}
\end{table*}

\begin{table*}[htbp]
\caption{The score for audio dialogue understanding performances under 16 different LALMs on ADU-Ambiguity dataset. The textual reference is \textit{before} the textual response in the evaluation prompt for the GPT-4 evaluator.
}
\small
\vspace{-1em}
\label{tab: app_adu_ambiguity_before}
\centering
\setlength{\tabcolsep}{10.8pt}{
\begin{tabular}{lc|cccc}
\toprule
\multicolumn{1}{l}{\multirow{2}{*}{Models}} & \multicolumn{1}{l|}{\multirow{2}{*}{Size}} & \multicolumn{4}{c}{ADU-Ambiguity} \\
&  & Intonation-based & Pause-based &  Homophone-based  & Repetition-based \\
\midrule 
PandaGPT & 7B & 0 & 0 & 0.98 & 0.98  \\
NExT-GPT & 7B & 0 & 0.01 & 0.99 & 0.99 \\
Qwen-Audio & 7B & 0 & 0.04 & 1.13 & 1.03 \\
Mini-Omni & 0.5B & 0.07 & 1.26 & 1.34 & 3.38 \\
SALMONN & 7B &  0.08 & 1.31 & 1.35 & 4.00 \\
Qwen-Audio-Chat & 7B & 0.01 & 0.52 & 1.70 & 5.18 \\
SpeechGPT & 7B & 0.13 & 1.19 & 2.70 & 4.70 \\
Moshi & 7B & 1.37 & 3.15 & 2.94 & 3.61 \\
SALMONN & 13B & 0.14 & 0.40 & 2.41 & 4.26 \\
BLSP & 7B & 2.01 & 2.92 & 2.38 & 5.70  \\
Step-Audio-Chat & 130B & 3.17 & 5.34 & 3.25 & 6.87 \\
\midrule
Whisper+LLaMA-2 & 7B & 3.02 & 3.65 & 2.65 & 7.86 \\
Whisper+LLaMA-3 & 8B & 3.40 & 4.44 & 2.76 & 8.56 \\
Whisper+LLaMA-3 & 70B & 3.64 & 4.56 & 2.92 & 8.76  \\
Whisper+GPT-4 & - & 4.02 & 5.02 & 3.64 & 9.03 \\
\midrule
GPT-4o & - & 6.96 & 5.11 & 5.97 & 9.10 \\
\bottomrule
\end{tabular}}
\end{table*}

\begin{table*}[htbp]
\caption{The score for audio dialogue understanding performances under 16 different LALMs on ADU-Ambiguity dataset. The textual reference is \textit{after} the textual response in the evaluation prompt for the GPT-4 evaluator.
}
\small
\vspace{-1em}
\label{tab: app_adu_ambiguity_after}
\centering
\setlength{\tabcolsep}{10.8pt}{
\begin{tabular}{lc|cccc}
\toprule
\multicolumn{1}{l}{\multirow{2}{*}{Models}} & \multicolumn{1}{l|}{\multirow{2}{*}{Size}} & \multicolumn{4}{c}{ADU-Ambiguity} \\
&  & Intonation-based & Pause-based &  Homophone-based  & Repetition-based  \\
\midrule 
PandaGPT & 7B & 0 & 0 & 0.99 & 1.00 \\
NExT-GPT & 7B & 0 & 0.02 & 1.10 & 1.00 \\
Qwen-Audio & 7B & 0 & 0.02 & 1.27 & 1.38 \\
Mini-Omni & 0.5B & 1.00 & 1.35 & 1.46 & 3.53 \\
SALMONN & 7B & 0.08 & 1.42 & 1.46 & 4.10  \\
Qwen-Audio-Chat & 7B & 0.02 & 0.57 & 1.94 & 5.48  \\
SpeechGPT & 7B & 0.15 & 1.30 & 2.82 & 5.00 \\
Moshi & 7B & 1.42 & 3.21 & 3.12 & 3.72 \\
SALMONN & 13B & 0.16 & 0.52 & 2.62 & 4.35 \\
BLSP & 7B & 2.22 & 3.23 & 2.39 & 6.09  \\
Step-Audio-Chat & 130B & 3.26 & 5.47 & 3.45 & 6.90 \\
\midrule
Whisper+LLaMA-2 & 7B & 3.27 & 3.88 & 2.75 & 8.02 \\
Whisper+LLaMA-3 & 8B & 3.98 & 4.66 & 2.86 & 8.64 \\
Whisper+LLaMA-3 & 70B & 4.23 & 4.87 & 3.26 & 8.81  \\
Whisper+GPT-4 & - & 4.35 & 5.20 & 3.86 & 9.14 \\
\midrule
GPT-4o & - & 7.11 & 5.32 & 6.12 & 9.20 \\
\bottomrule
\end{tabular}}
\end{table*}

\end{document}